\documentclass[runningheads,a4paper]{llncs}

\usepackage{amssymb}
\usepackage{graphicx}
\usepackage{color}
\usepackage{url}
\usepackage[table]{xcolor}
\usepackage{multirow}
\usepackage{subfig}

\newcommand\blfootnote[1]{%
  \begingroup
  \renewcommand\thefootnote{}\footnote{#1}%
  \addtocounter{footnote}{-1}%
  \endgroup
}

\urldef{\mail}\path|{j.m.wolterink, t.leiner, m.viergever, i.isgum}@umcutrecht.nl|

\newcommand{\keywords}[1]{\par\addvspace\baselineskip
\noindent\keywordname\enspace\ignorespaces#1}

\usepackage{array}
\newcolumntype{L}[1]{>{\raggedright\let\newline\\\arraybackslash\hspace{0pt}}m{#1}}
\newcolumntype{C}[1]{>{\centering\let\newline\\\arraybackslash\hspace{0pt}}m{#1}}
\newcolumntype{R}[1]{>{\raggedleft\let\newline\\\arraybackslash\hspace{0pt}}m{#1}}

\begin{document}

\mainmatter  

\title{Dilated Convolutional Neural Networks for Cardiovascular MR Segmentation in \\ Congenital Heart Disease}
\titlerunning{Dilated Convolutional Neural Networks for Cardiac MR Segmentation}

\author{Jelmer M. Wolterink\inst{1}, Tim Leiner\inst{2}, Max A. Viergever\inst{1}, Ivana I\v{s}gum\inst{1}}

\authorrunning{J.M. Wolterink et al.}

\institute{Image Sciences Institute, University Medical Center Utrecht, The Netherlands \and Department of Radiology, University Medical Center Utrecht, The Netherlands}
\maketitle

\begin{abstract}
We propose an automatic method using dilated convolutional neural networks (CNNs) for segmentation of the myocardium and blood pool in cardiovascular MR (CMR) of patients with congenital heart disease (CHD).

Ten training and ten test CMR scans cropped to an ROI around the heart were provided in the MICCAI 2016 HVSMR challenge. A dilated CNN with a receptive field of $131\times 131$ voxels was trained for myocardium and blood pool segmentation in axial, sagittal and coronal image slices. Performance was evaluated within the HVSMR challenge. 

Automatic segmentation of the test scans resulted in Dice indices of $0.80\pm0.06$ and $0.93\pm0.02$, average distances to boundaries of $0.96\pm 0.31$ and $0.89\pm 0.24$ mm, and Hausdorff distances of $6.13\pm 3.76$ and $7.07\pm 3.01$ mm for the myocardium and blood pool, respectively. Segmentation took $41.5\pm 14.7$ s per scan. 

In conclusion, dilated CNNs trained on a small set of CMR images of CHD patients showing large anatomical variability provide accurate myocardium and blood pool segmentations.
\end{abstract}

\blfootnote{This work has been published as: Wolterink J.M, Leiner T., Viergever M.A., I\v{s}gum I. (2017) Dilated Convolutional Neural Networks for Cardiovascular MR Segmentation in Congenital Heart Disease. In: \textit{Reconstruction, Segmentation, and Analysis of Medical Images. RAMBO 2016, HVSMR 2016}, Lecture Notes in Computer Science, vol 10129, pp. 95--102}

\section{Introduction}
Congenital heart diseases (CHD) are a type of congenital defect affecting almost 1\% of live births \cite{Gilb16}. 
Patients with severe congenital heart disease often require surgery in their childhood. It has been shown that the use of patient-specific 3D models is helpful for preoperative planning \cite{Schm15}. Such models are typically based on a segmentation of the patient's anatomy in cardiovascular MR (CMR). 
However, segmentation of cardiac structures in CMR requires several hours of manual annotations \cite{Valv15}. Hence, there is a need for semi-automatic or fully automatic segmentation methods to speed up this time-consuming process and reduce the workload for clinicians.

\begin{figure}[t!]
\includegraphics[height=0.20\textwidth]{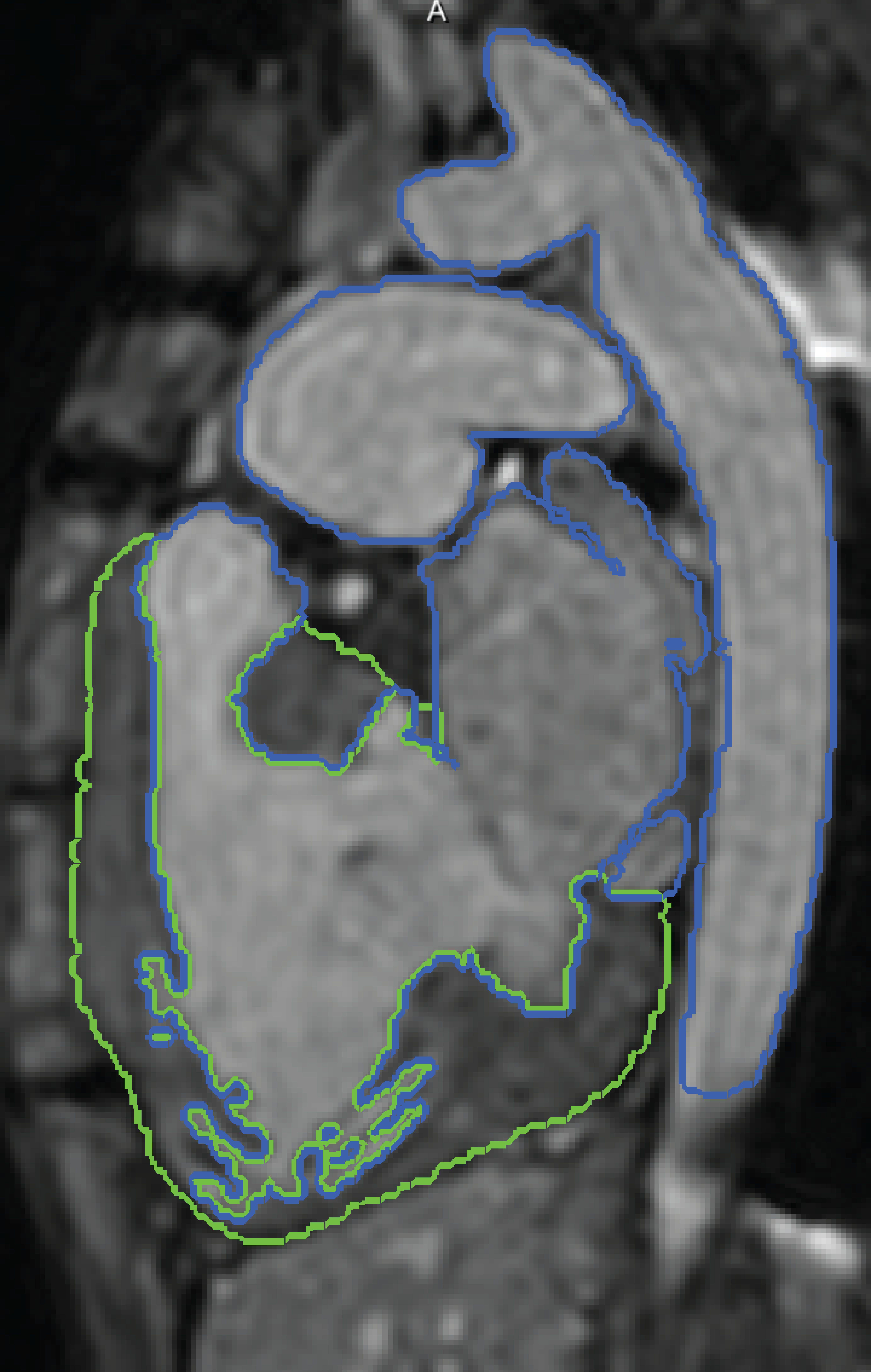} \hfill
\includegraphics[height=0.20\textwidth]{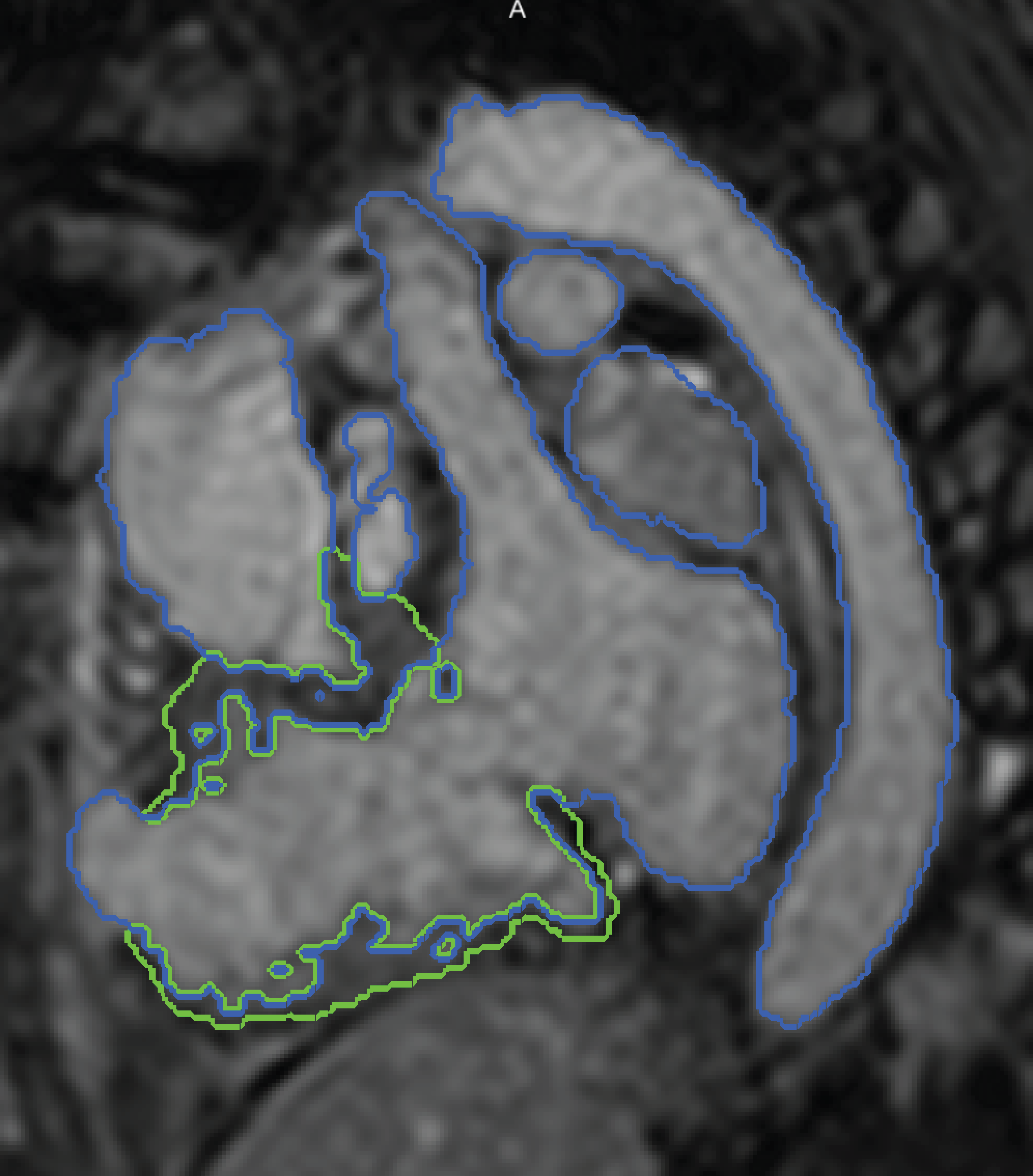} \hfill
\includegraphics[height=0.20\textwidth]{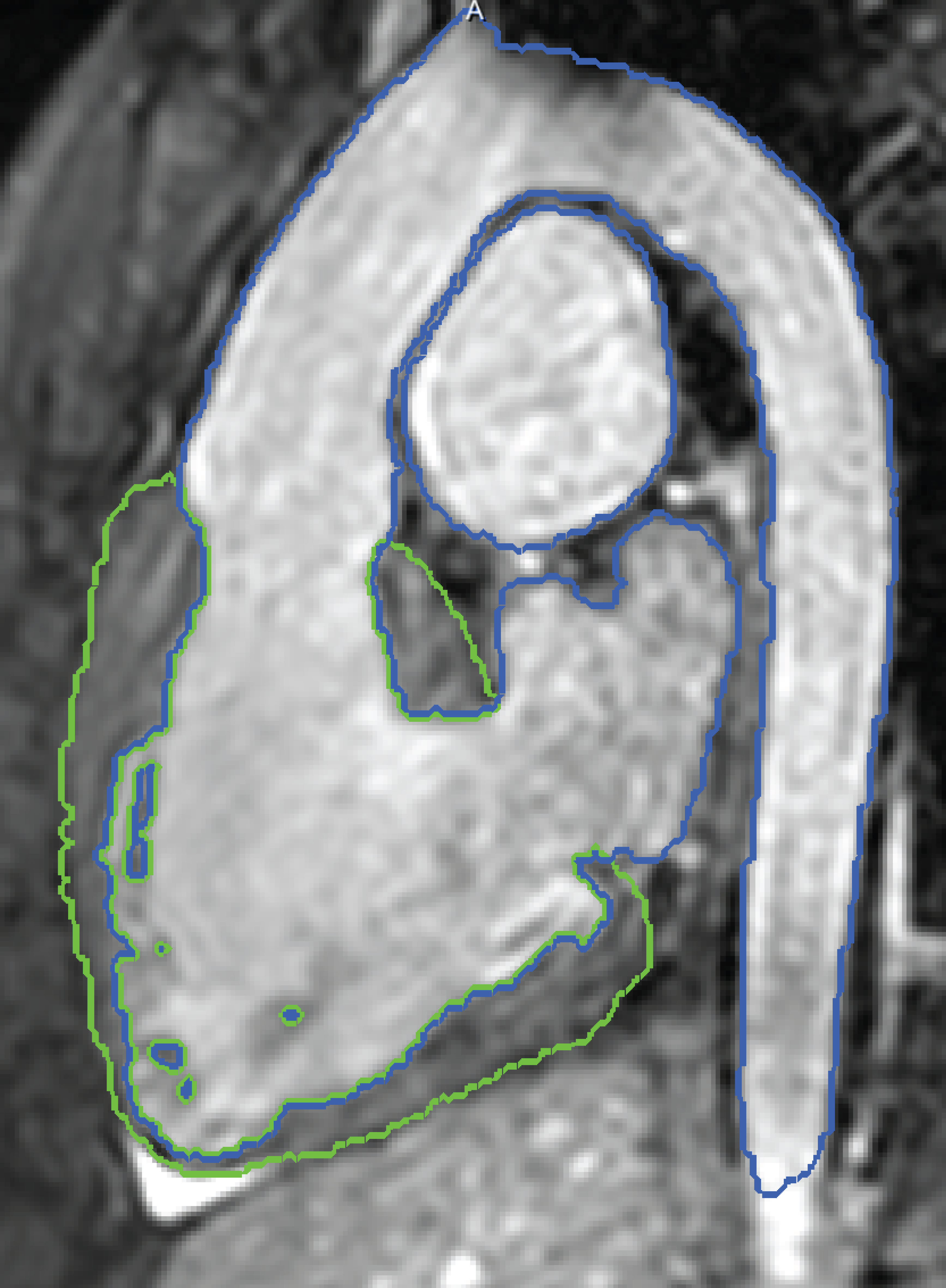} \hfill
\includegraphics[height=0.20\textwidth]{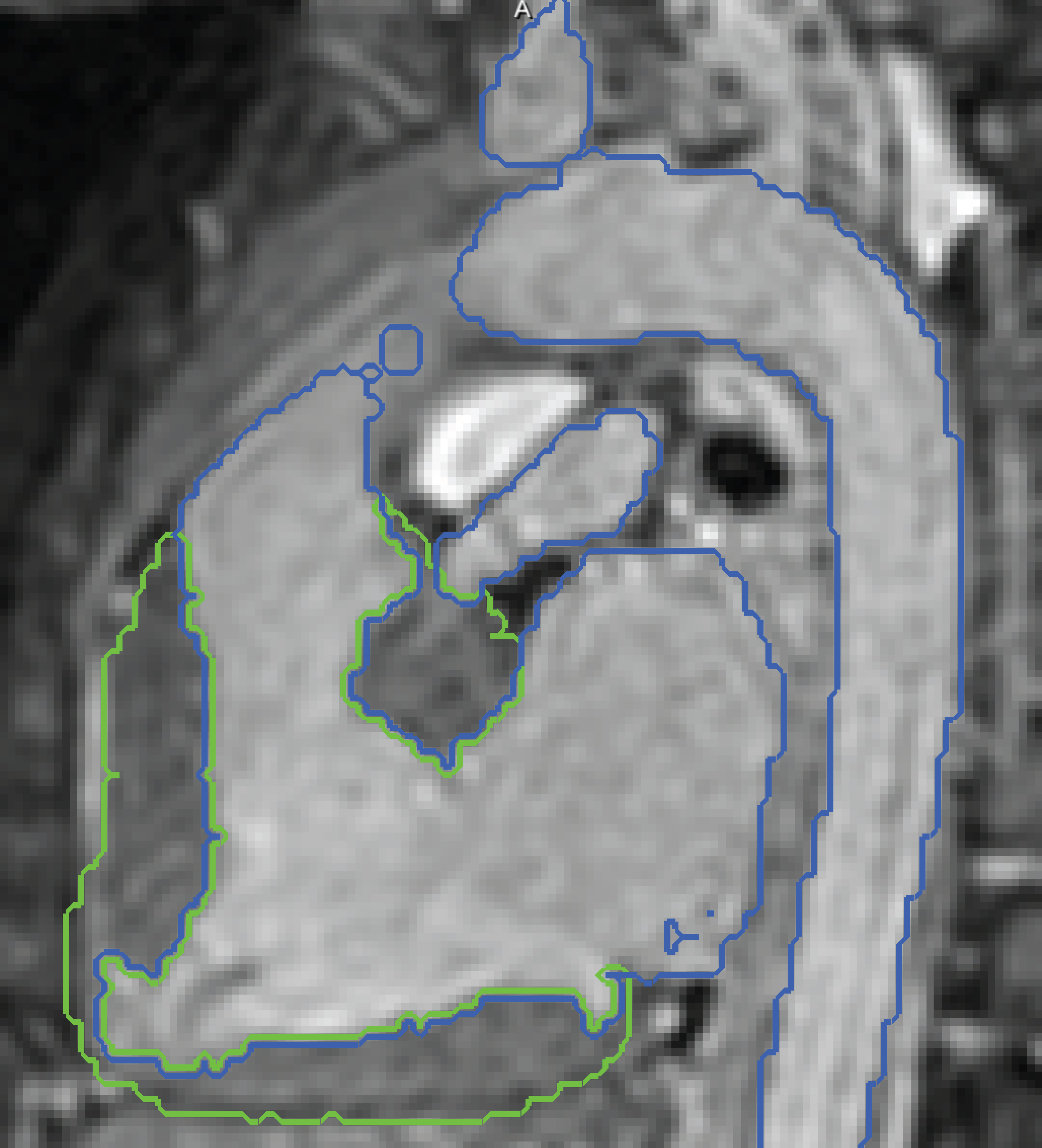} \hfill
\caption{Example cardiovascular MR data of four patients with congenital heart disease. The examples illustrate the high variability in the structure and appearance of the blood pool and myocardium. Reference annotations for the blood pool and myocardium are shown in blue and green, respectively.}
\label{fig:example}
\end{figure}

The large anatomical variability among patients poses a major challenge for (semi)automatic segmentation of CMR in CHD patients (Fig. \ref{fig:example}). Methods relying on multi-atlas based segmentation would require a highly diverse training set representing the various manifestations of CHD. Hence, local analysis based on intensity and texture might be advantageous. Pace et al. proposed a patch-based semi-automatic segmentation method that produces highly accurate segmentations, requiring one hour of manual interaction and one hour of offline processing per scan \cite{Pace15}. The label of each voxel in an image is determined based on patch similarity to manually segmented sections in the image. To eliminate any user interaction, we propose a fully automatic patch-based voxel classification method. Voxel labels are determined based on similarities to voxels in training images using a convolutional neural network (CNN). 

CNNs have been widely adopted in medical image analysis for segmentation of e.g. tissue classes \cite{Moes16} and tumors \cite{Hava16} in brain MR, neuronal structures in electron-miscroscopy \cite{Ronn15} and coronary artery calcium in cardiac CT angiography \cite{Wolt16}. 
A CNN labels each voxel in an image based on one or multiple patches surrounding that voxel. An effective voxel classification method should combine both local structure information and global context information. To this end, Moeskops et al. proposed a multi-scale approach using differently-sized patches extracted for every voxel \cite{Moes16}, and Ronneberger et al. used a CNN which merges information at different scales by skipping layers \cite{Ronn15}. 
However, patch extraction at every voxel is time-consuming and downsampling layers affect the output resolution and translational equivariance, meaning that the exact output may depend on the positioning of the input. 

Recently, stacks of dilated convolutions have been proposed for image segmentation \cite{Yu15}. Such stacks aggregate features at multiple scales through convolutional layers with very few parameters, thereby avoiding problems such as overfitting, while generating high resolution output images with translation equivariance. The promise of a large receptive field with few trainable parameters is particularly interesting in medical imaging, where data sets are often small. In this work, we use CNNs with dilated convolutions to automatically segment CMR images of CHD patients.

\section{Data}
The method was developed and evaluated within the framework of the MICCAI Workshop on Whole-Heart and Great Vessel Segmentation from 3D Cardiovascular MRI in Congenital Heart Disease (HVSMR 2016)\footnote{\url{http://segchd.csail.mit.edu}}. Ten training and ten test CMR scans were provided by the workshop organizers. 
The  scans were acquired at Boston Children's Hospital with a 1.5T Philips Achieva scanner (TR = 3.4 ms, TE = 1.7 ms, $\alpha$ = 60$^{\circ}$). Three images were provided for each patient: a complete axial CMR image, the same image cropped around the heart and thoracic aorta, and a cropped short axis reconstruction. In the current work, we use the cropped image around the heart and thoracic aorta. Reconstructed in-plane resolution of the images ranged from 0.73 mm to 1.15 mm, while slice spacing ranged from 0.65 mm to 1.15 mm. 

Reference segmentations of the blood pool and myocardium were provided for the training scans, but not for the test scans. These segmentations were made by a trained observer and validated by two clinical experts. The blood pool class contained both atria and ventricles, the aorta, pulmonary veins, and superior and inferior vena cava. The myocardium class contained the thick muscle around the two ventricles and their separating septum. Example reference segmentations are shown in Fig. \ref{fig:example}.

\section{Methods}
We trained a purely convolutional CNN to assign a class label to each voxel in a CMR volume based on classification of three orthogonal patches centered at the voxel. The CNN uses dilated convolutions allowing large receptive fields with few trainable parameters.

\begin{figure}[t!]
\centering
\includegraphics[width=0.9\textwidth]{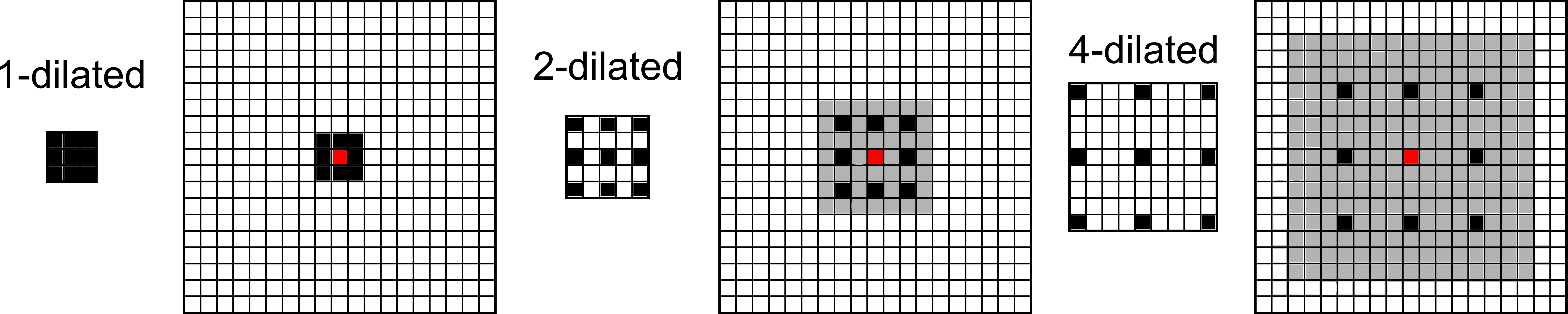}
\caption{Convolution with a standard 1-dilated 3$\times$3 kernel, followed by convolution with a 2-dilated and a 4-dilated kernel (kernels shown in black). The receptive field (shown in gray) increases from $3\times 3$ after the first convolution, to $7\times 7$ after the second convolution and $15\times 15$ after the third convolution. All convolutions only use $3\times 3=9$ trainable parameters.}
\label{fig:kernels}
\end{figure}

CNNs consist of a sequence of convolution layers, which convolve an image $F_l$ at layer $l$ with a kernel $k$ to obtain image $F_{l+1}$ at layer $l+1$. Dilated convolutions are extensions of these convolutions, that add spacing between the elements of the kernel $k$ so that neighboring voxels at larger intervals are considered when computing the value for a voxel $x$ in $F_{l+1}$. The level of dilation determines the stride between kernel elements (Fig. \ref{fig:kernels}).  CNNs with dilated convolutions have several advantages over CNNs with non-dilated, i.e. standard, convolutions. First, by stacking convolution layers with increasing levels of dilation, the receptive field for every voxel can be substantially extended at the cost of only few additional trainable parameters. 
Second, dilated convolution operations are translationally equivalent: the same multi-scale feature aggregation pyramid is applied at each location in the image. Hence, translating the image results in a translated version of the original output. Third, no downsampling layers are required to obtain large receptive fields and hence, high resolution label maps can be directly predicted by the network.

\begin{table}[t!]
\caption{The convolutional neural network architecture used in this study. For each layer, the convolution kernel size, the level of dilation, the receptive field, the number of output channels and the number of trainable parameters are listed. Figures in the top row illustrate the receptive field at each layer shown in red.}
\label{tab:layers}
\resizebox{\textwidth}{!}{  
\scriptsize{
\begin{tabular}{|l|C{1.15cm}|C{1.15cm}|C{1.15cm}|C{1.15cm}|C{1.15cm}|C{1.15cm}|C{1.15cm}|C{1.15cm}|C{1.15cm}|C{1.15cm}|}
\multicolumn{1}{c|}{}						&	\includegraphics[width=1.10cm]{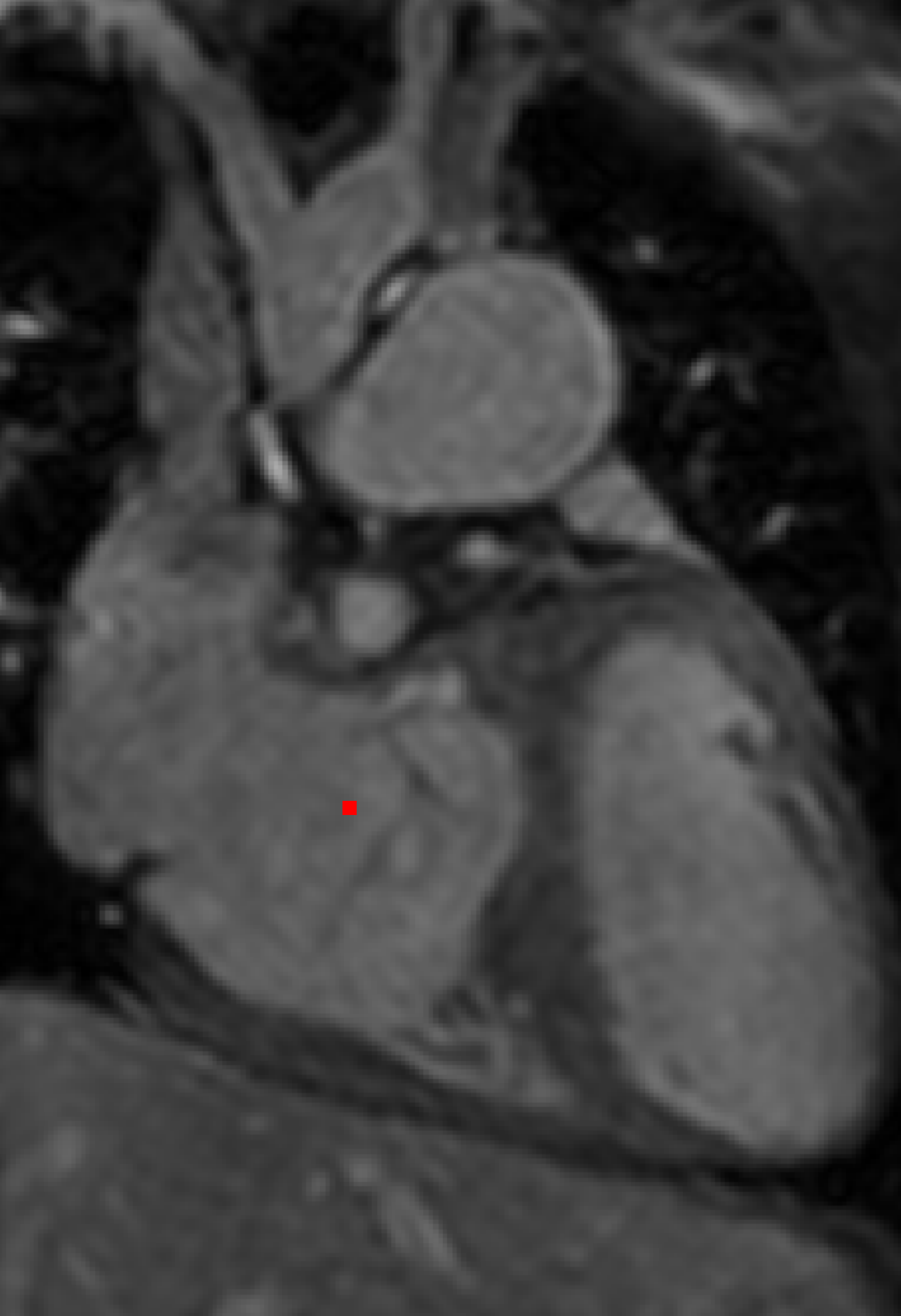} &	\includegraphics[width=1.10cm]{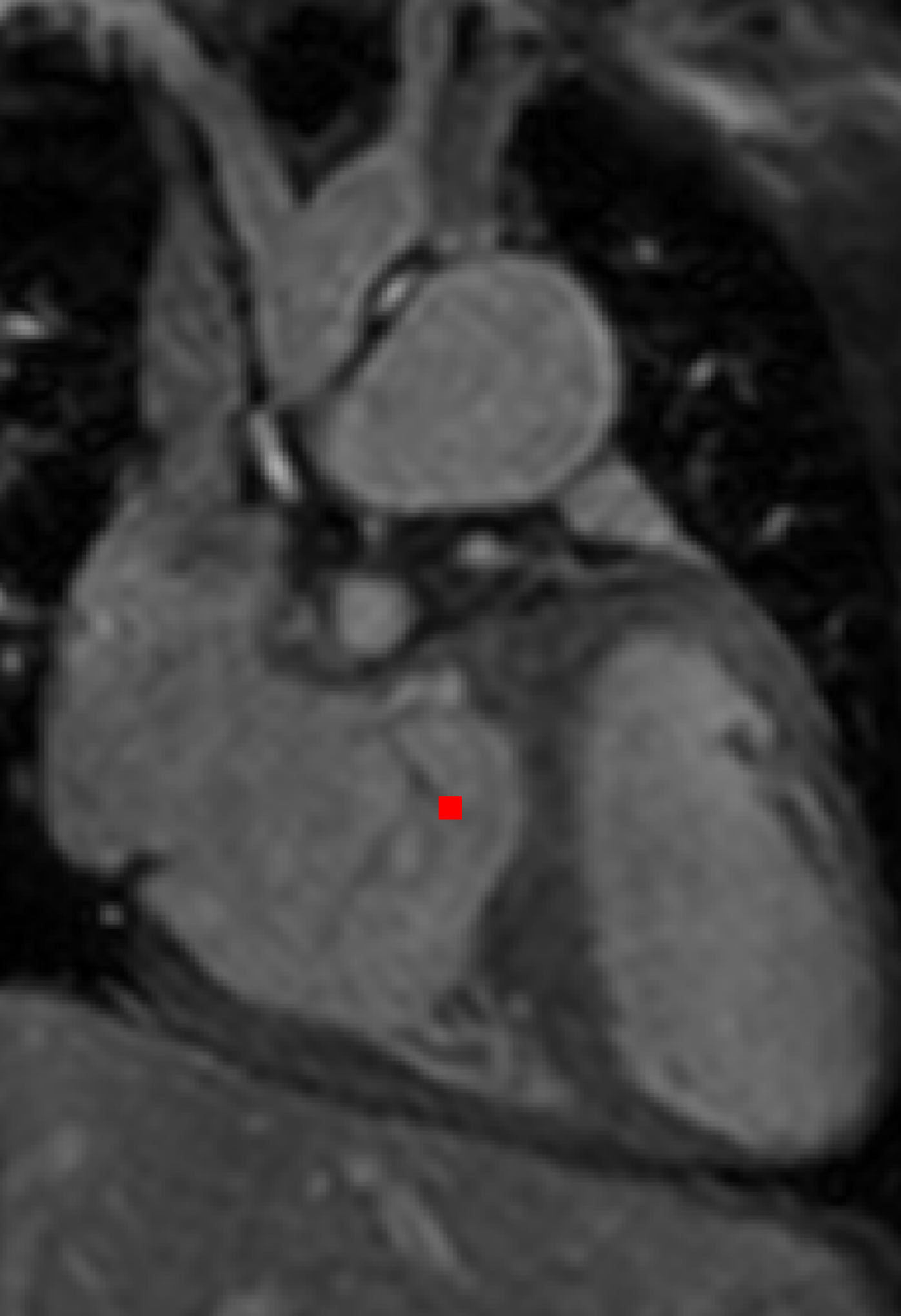} &	\includegraphics[width=1.10cm]{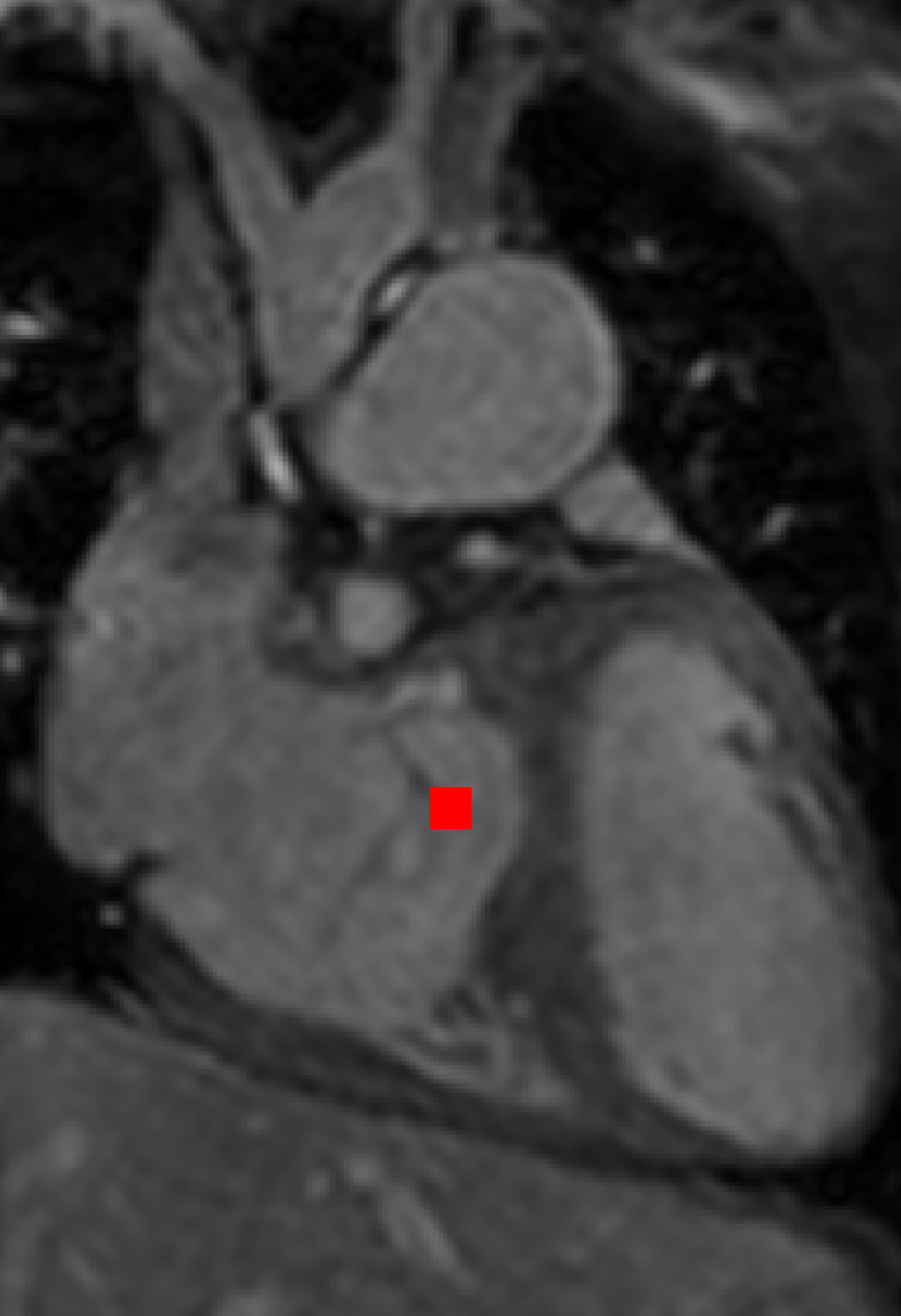} &	\includegraphics[width=1.10cm]{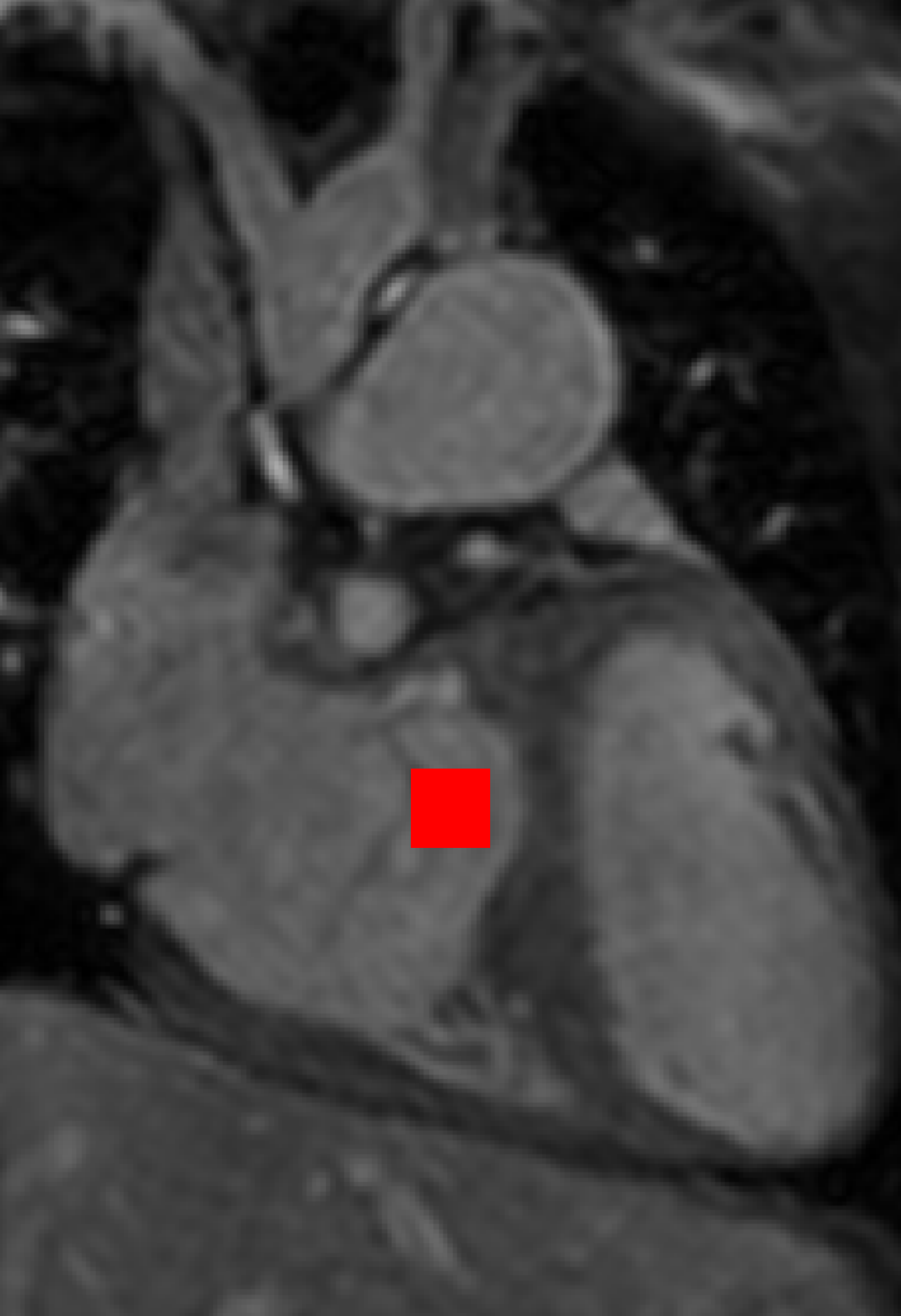} &	\includegraphics[width=1.10cm]{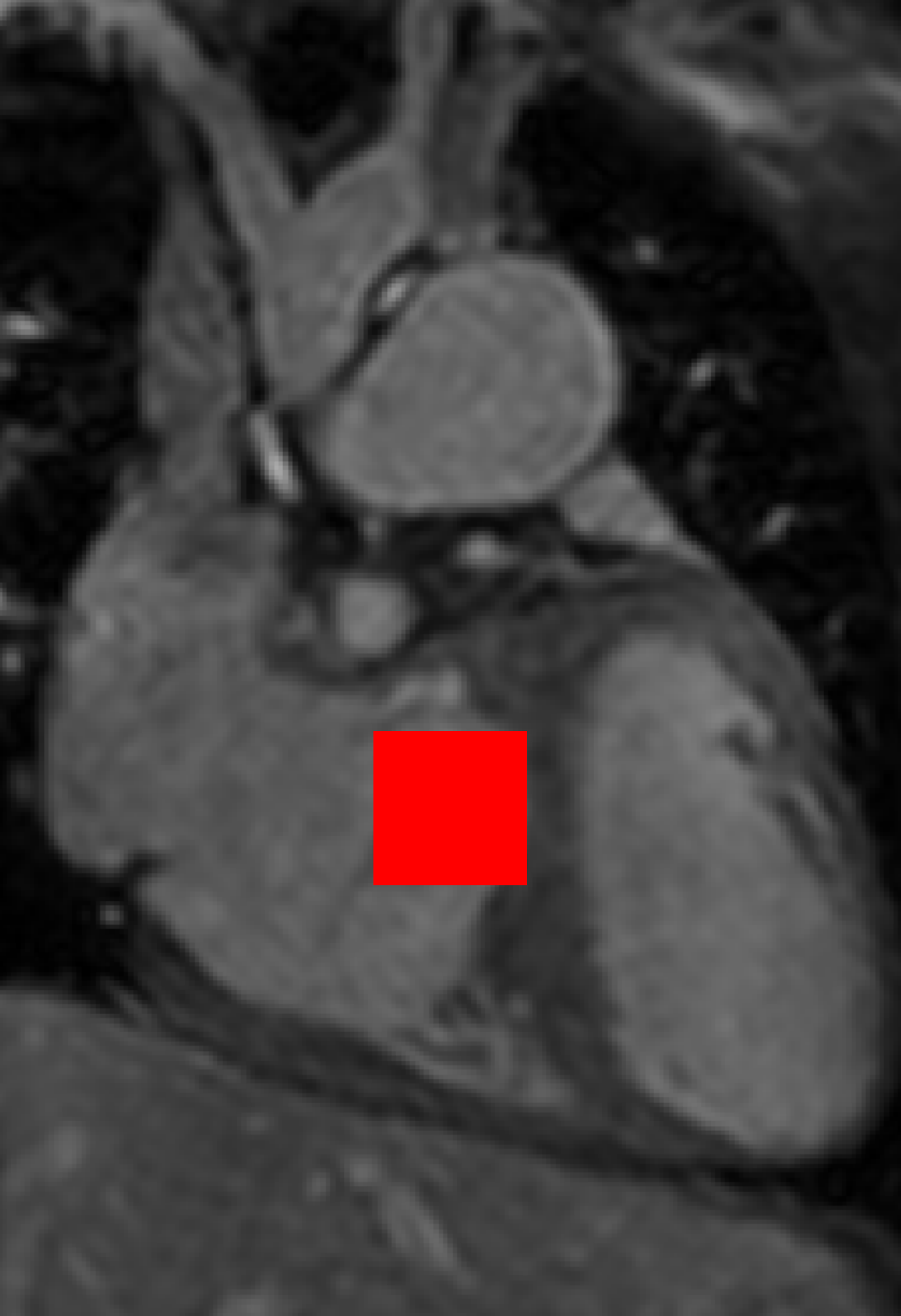} &	\includegraphics[width=1.10cm]{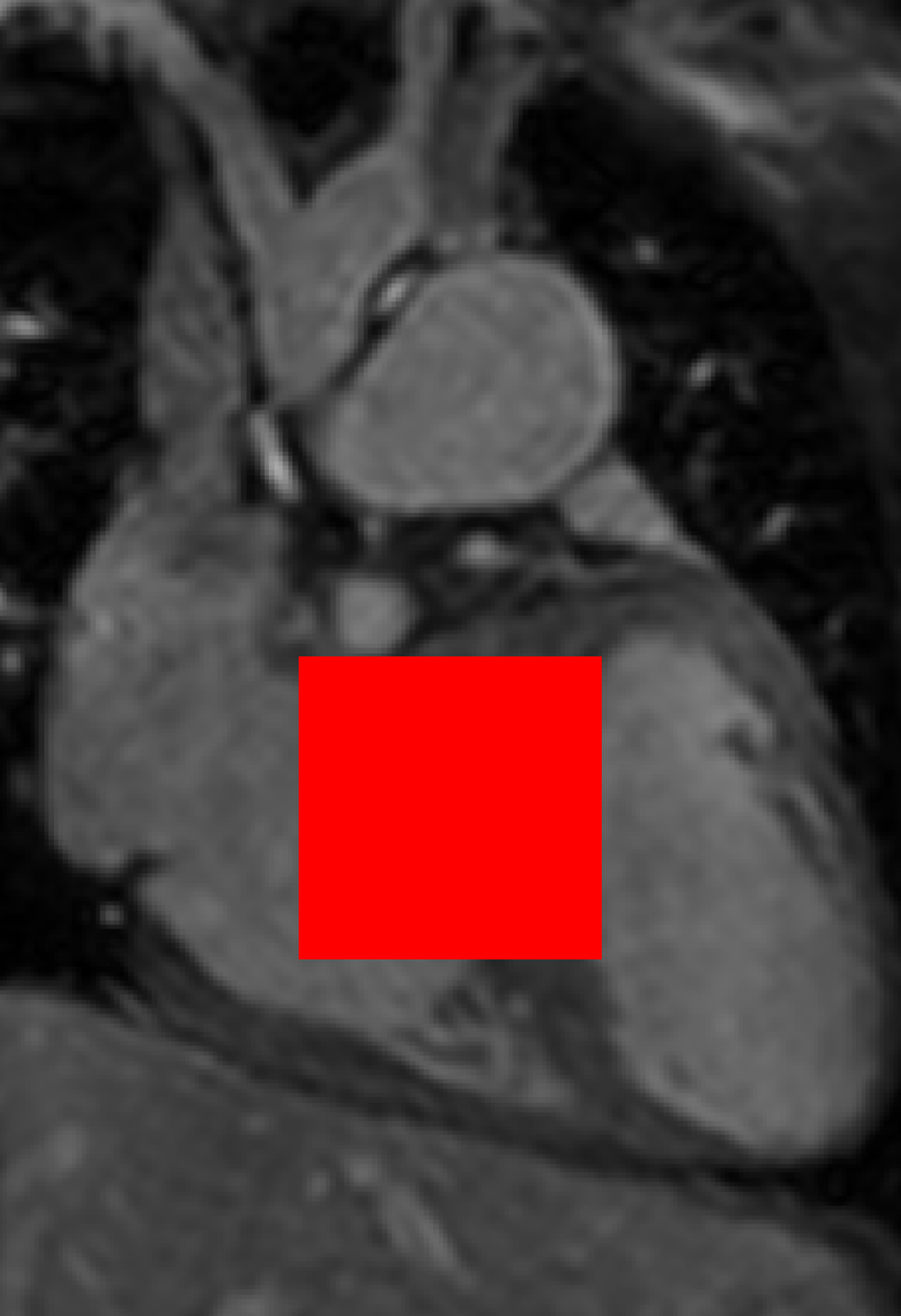} &	\includegraphics[width=1.10cm]{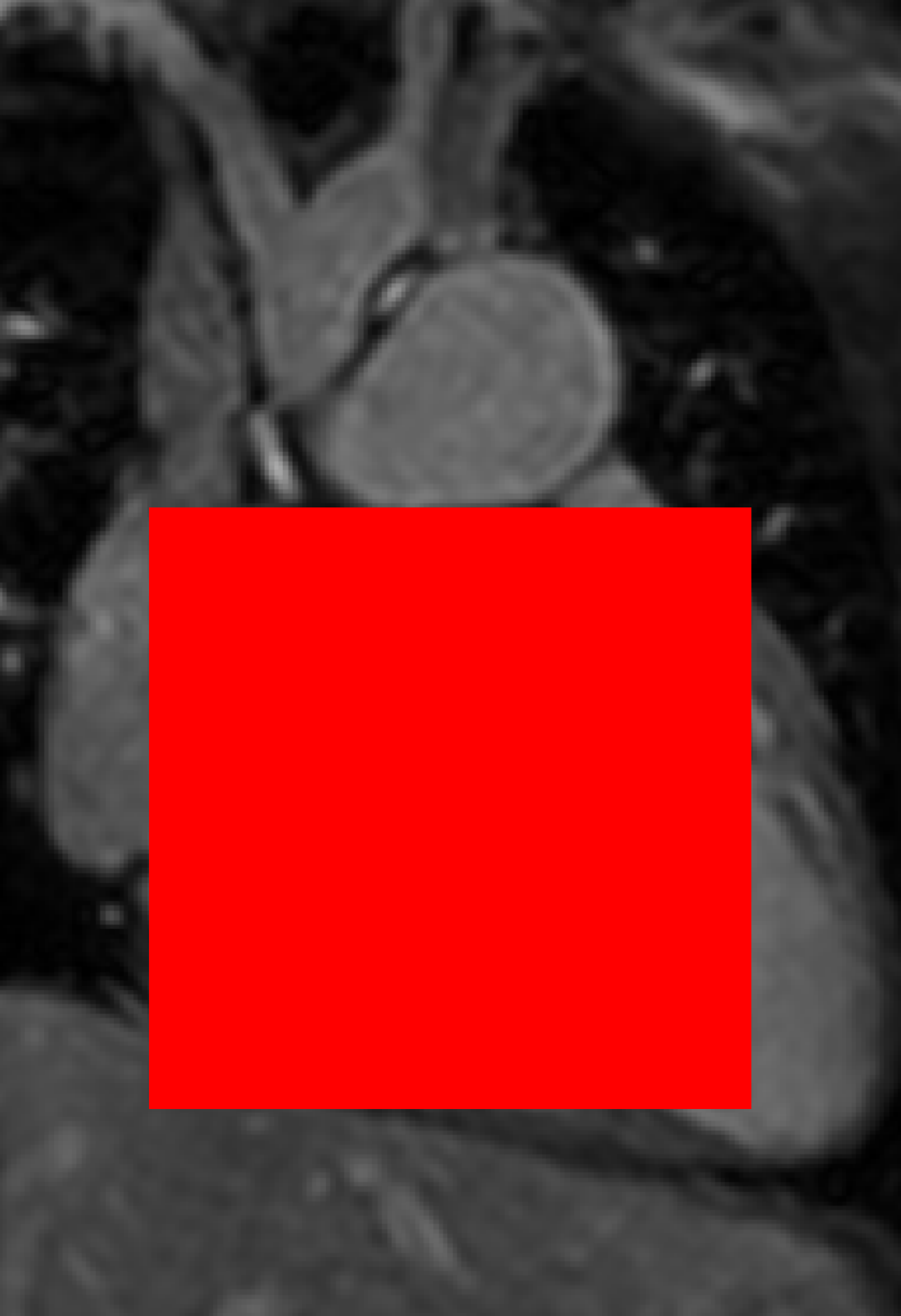} &	\includegraphics[width=1.10cm]{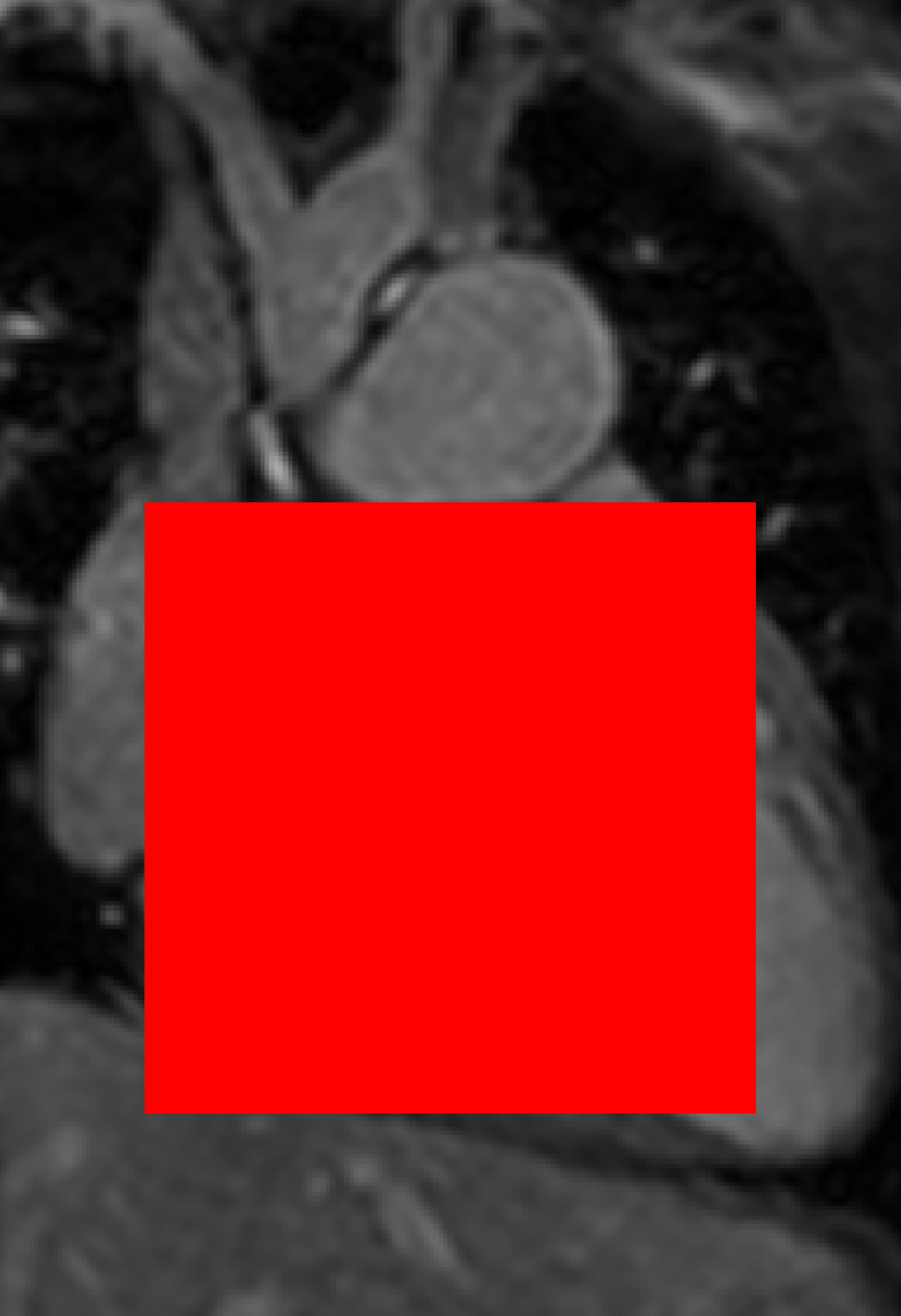} & \includegraphics[width=1.10cm]{field131.pdf} & \includegraphics[width=1.10cm]{field131.pdf} \\ \hline
Layer							&	1 					&	2 					&	3 					&	4 						&	5 						&	6 						&	7 							&	8 							& 9 							& 10 \\ \hline
Convolution				&	3$\times$3  & 3$\times$3	&	3$\times$3 	& 3$\times$3		&	3$\times$3 		&	3$\times$3 		& 3$\times$3 			&	3$\times$3 			& 1$\times$1 			& 1$\times$1	\\ \hline
Dilation					& 1						& 1						& 2						& 4							& 8							& 16						& 32							& 1 							& 1								& 1				\\ \hline
Field							&	3$\times$3 	& 5$\times$5	&	9$\times$9	&	17$\times$17 &	33$\times$33	&	65$\times$65 	&	129$\times$129 	& 131$\times$131 	& 131$\times$131 	& 131$\times$131 \\ \hline
Channels	   			&	32 					& 32 					&	32 					& 32 						&	32  					&	32	 	 		  	&	32							&	32							& 192							& 3					\\	\hline 
Parameters						& 320					& 9248				& 9248				& 9248					& 9248					& 9248					& 9248						& 9344						& 6912						& 579				\\ \hline
\end{tabular} 
}
}
\end{table}

Table \ref{tab:layers} lists the CNN architecture used in this study. We adapt the dilated convolution context module proposed by Yu et al. \cite{Yu15} and extend the receptive field from $67\times 67$ voxels to $131 \times 131$ voxels by inserting layer 7, with 32-dilated kernels. Layers 1 to 8 serve as feature extraction layers, while layers 9 and 10 are fully connected classification layers, implemented as $1\times 1$ convolutional layers for increased efficiency. In each feature extraction layer, 32 kernels are used. The dilation level is increased between layers 2 and 7. This allows the receptive field to grow exponentially, while the number of trainable parameters grows linearly; the same number of parameters is used in each layer. The figures in the top row illustrate the receptive field in each layer. Layers 1 to 9 are each followed by an exponential linear unit (ELU) activation function \cite{Clev16}, while layer 10 is followed by a softmax function. Batch normalization and dropout are applied to the fully connected layers 8 and 9 \cite{Ioff15,Sriv14}. Classification is performed in a wider layer with 192 channels and a final layer with 3 output channels, i.e. for the myocardium, blood pool and background. In total, the network contains 72,643 trainable parameters.

To correct for differences in intensity signals between CMR images, each image was normalized to have zero mean and unit variance. Furthermore, to correct for potential differences in orientation of patients, copies of the images rotated 90, 180 and 270 degrees along each axis were added to the training set. Finally, to guarantee that structures appear at similar scales in different images and along different axes, all images were resampled to an isotropic resolution of 0.65$\times$0.65$\times$0.65 mm per voxel, the smallest voxel dimension present in the data set. These isotropic volumes were used for voxel classification. 
A single CNN was trained to segment axial, sagittal and coronal image slices. During testing, full slices with 65-voxel zero-padding were processed so that for each viewing direction, 3D probabilistic maps were obtained for the myocardium, blood pool and background classes. These maps were averaged per segmentation class and resampled to the original input dimensions using trilinear interpolation. Finally, each voxel was assigned the segmentation class label with the highest posterior probability. Hence, the final probability for the myocardium, blood pool or background for each voxel depends on three $131\times 131$ patches centered at that voxel. To guarantee contiguous myocardium and blood pool segmentations, only the largest component for each class was included in the final segmentation.

Evaluation was performed through an online system provided by the HVSMR challenge. The overlap between reference and automatically obtained segmentations was computed using the Dice index. Furthermore, the difference between reference and automatically obtained boundaries was computed using the average distance to boundaries (ADB) and the Hausdorff distance.

\begin{table*}[t!]
\centering
\caption{Results for the training and test set as provided by the HVSMR challenge. For both the myocardium and the blood pool, the Dice index, the average distance to boundaries (ADB) and the Hausdorff distance (Hausdorff) are listed.}
\label{tab:results}
\scriptsize
\begin{tabular}{ll| L{1.5cm} L{1.5cm} L{1.5cm} | L{1.5cm} L{1.5cm} L{1.5cm}}
        &            & \multicolumn{3}{l|}{Myocardium} & \multicolumn{3}{l}{Blood pool} \\
        &            & Dice    & ADB     & Hausdorff   & Dice    & ADB     & Hausdorff   \\ \hline
Training & Average & $0.80 \pm 0.06$    & $1.01 \pm 0.43$    & $6.70 \pm 3.52$        & $0.92\pm 0.03$    & $0.81 \pm 0.28$   & $5.86\pm 3.36$ \\ \hline
Test & Patient 10 & 0.72 & 1.34 & 10.75 & 0.94 & 0.74 & 5.23  \\
     & Patient 11 & 0.81 & 0.68 & 2.50  & 0.93 & 0.94 & 9.17  \\
     & Patient 12 & 0.87 & 0.60 & 3.94  & 0.93 & 0.83 & 9.74  \\
     & Patient 13 & 0.88 & 1.03 & 10.19 & 0.94 & 0.94 & 10.62 \\
     & Patient 14 & 0.71 & 1.33 & 8.69  & 0.90 & 1.07 & 4.21  \\
     & Patient 15 & 0.76 & 1.07 & 3.97  & 0.89 & 1.44 & 11.78 \\
     & Patient 16 & 0.76 & 0.80 & 3.14  & 0.91 & 0.77 & 6.12  \\
     & Patient 17 & 0.87 & 0.70 & 4.14  & 0.95 & 0.61 & 4.27  \\
     & Patient 18 & 0.85 & 0.61 & 2.19  & 0.94 & 0.64 & 3.29  \\
     & Patient 19 & 0.79 & 1.41 & 11.76 & 0.93 & 0.87 & 6.28  \\ 
     & Average      & $0.80 \pm 0.06$    & $0.96 \pm 0.32$    & $6.13 \pm 3.76$        & $0.93\pm 0.02$    & $0.89\pm 0.24$    & $7.07\pm 3.01$         \\
\end{tabular}
\end{table*}

\section{Experiments and Results}
We performed a five-fold cross-validation experiment on the training set, where each fold contained two CMR scans. Furthermore, to segment the test set, we trained a single CNN using all training images.
Network parameters were optimized with Adam \cite{King15} using categorical cross-entropy as the cost function. Each CNN was trained with 10,000 training steps, which required 12 hours using a state-of-the-art GPU. In each training step a mini-batch containing 128 randomly selected $201\times 201$ subimages from the training set was provided. Hence, in each training step the network optimized parameters for $71\times 71\times 128=645,248$ training voxels.

\begin{figure}[t!]
\subfloat[]{
\includegraphics[width=0.15\textwidth]{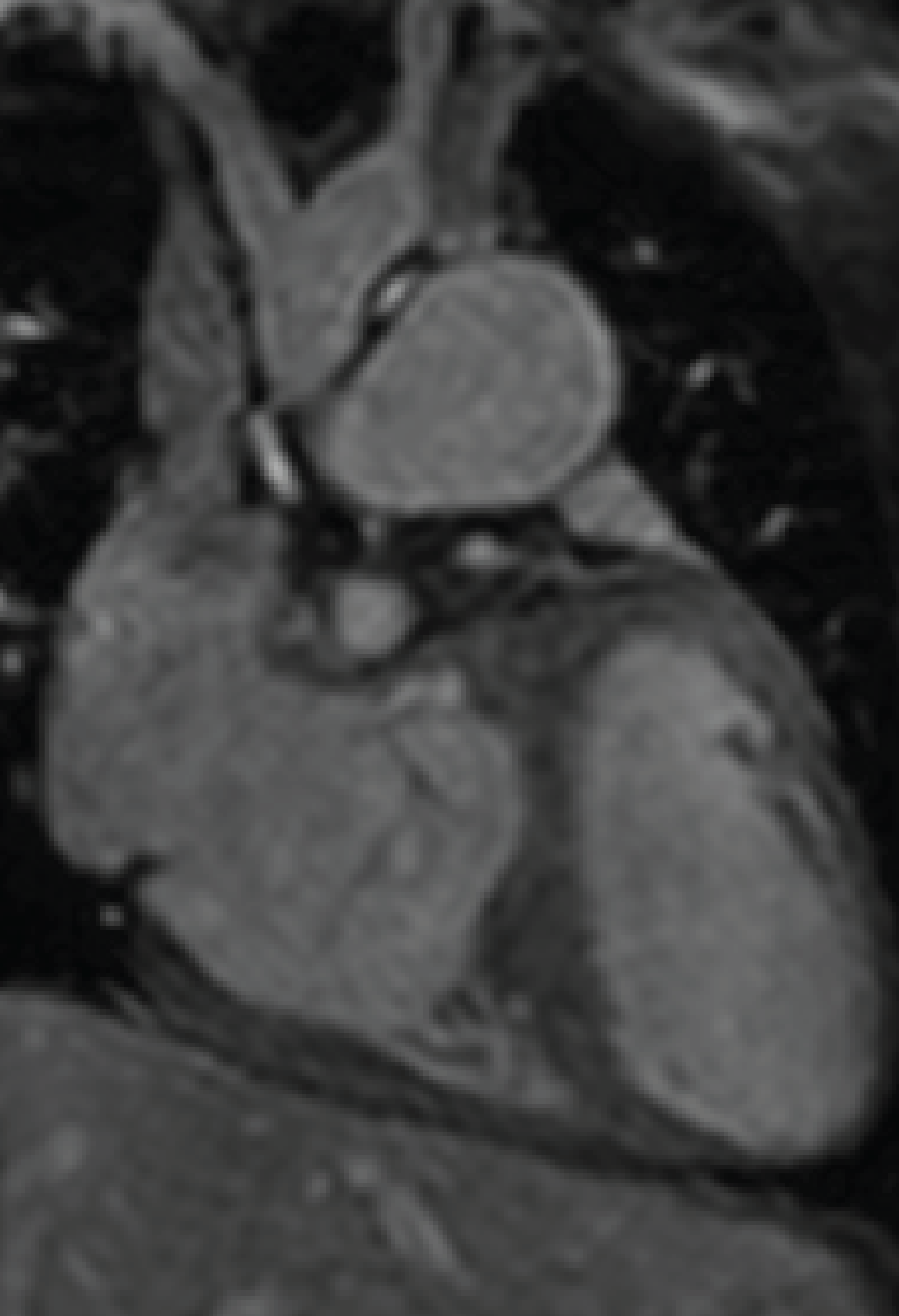}
\label{subfig:origslice}
}
\hfill
\subfloat[]{
\includegraphics[width=0.15\textwidth]{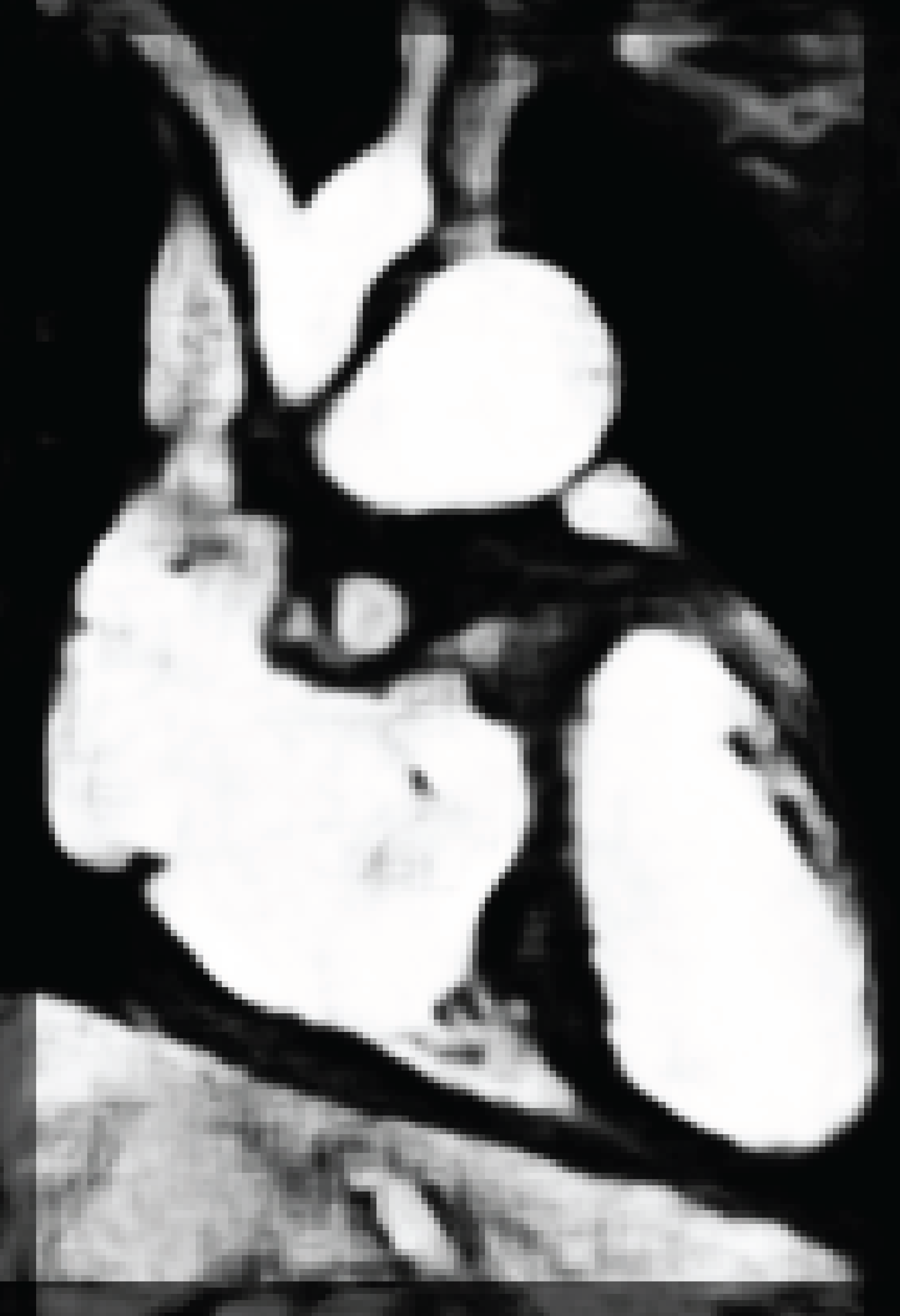}
\label{subfig:smallreceptive}
}
\hfill
\subfloat[]{
\includegraphics[width=0.15\textwidth]{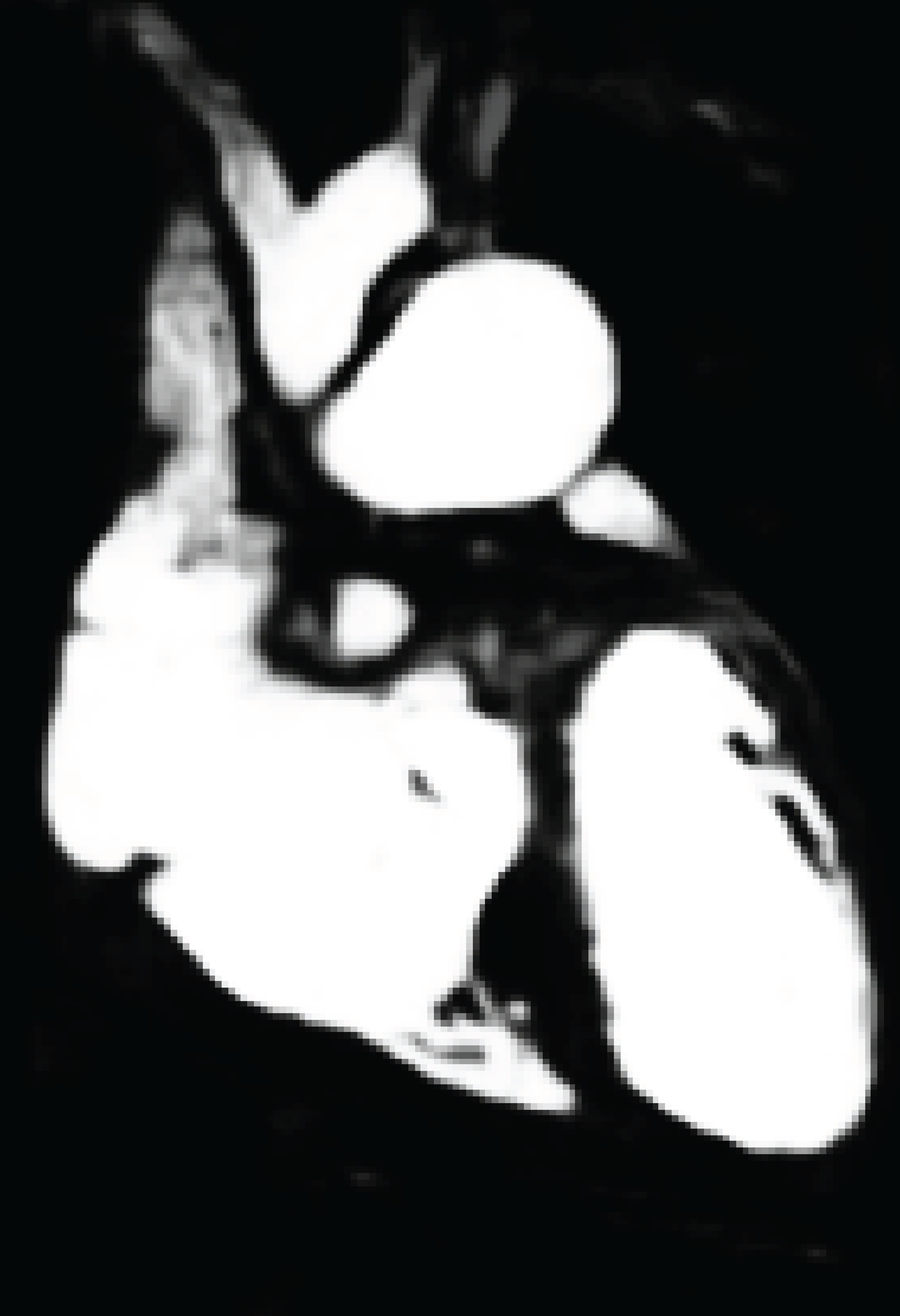}
\label{subfig:largereceptive}
}
\hfill
\subfloat[]{
\includegraphics[width=0.15\textwidth]{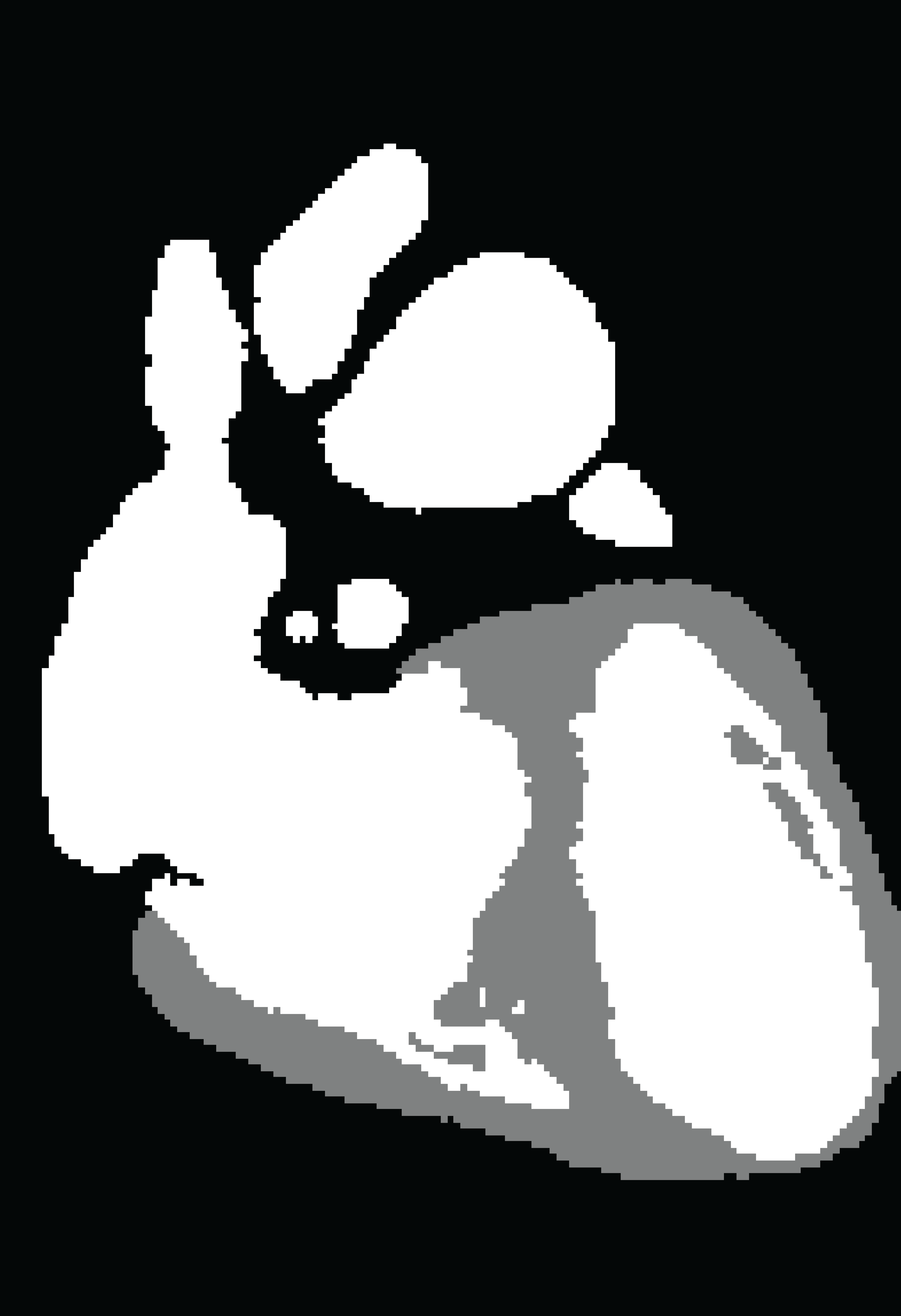}
\label{subfig:reference}
}
\caption{\protect\subref{subfig:origslice} Example CMR image slice. Probabilistic blood pool maps obtained using \protect\subref{subfig:smallreceptive} a CNN without dilation ($17\times 17$ voxel receptive field) showing oversegmentation in the liver and \protect\subref{subfig:largereceptive} a CNN with dilation ($131\times 131$ voxel receptive field) showing no response in the liver. \protect\subref{subfig:reference} reference annotation with the blood pool shown in white. Both networks have 72,643 trainable parameters.}
\label{fig:withwithout}
\end{figure}

Table \ref{tab:results} lists the Dice index, the average distance to boundary (ADB), and the Hausdorff distance for automatic myocardium and blood pool segmentation in each of the ten test scans provided by the HVSMR challenge, as well as average scores for the training and test sets. Automatic segmentation of the \textit{training} scans resulted in Dice indices of $0.80\pm0.06$ and $0.92\pm0.03$, average distances to boundaries of $1.01\pm 0.43$ and $0.81\pm 0.28$ mm, and Hausdorff distances of $6.70\pm 3.52$ and $5.86\pm 3.36$ mm for the myocardium and blood pool, respectively. 
Automatic segmentation of the \textit{test} scans resulted in Dice indices of $0.80\pm0.06$ and $0.93\pm0.02$, average distances to boundaries of $0.96\pm 0.31$ and $0.89\pm 0.24$ mm, and Hausdorff distances of $6.13\pm 3.76$ and $7.07\pm 3.01$ mm for the myocardium and blood pool, respectively. 
The Dice index for myocardium was in all cases lower than the Dice index for blood pool segmentation. In several cases, Dice indices for the blood pool were affected by the (partial) identification of vessels that were not included in the reference standard, as shown in Fig. \ref{subfig:largereceptive}. Segmentation of a CMR image took between 12.9 and 64.0 seconds, depending on image size, with an average of $41.5\pm 14.7$ seconds.

\begin{figure}[t!]
\hfill
\subfloat[]{
\includegraphics[width=0.12\textwidth]{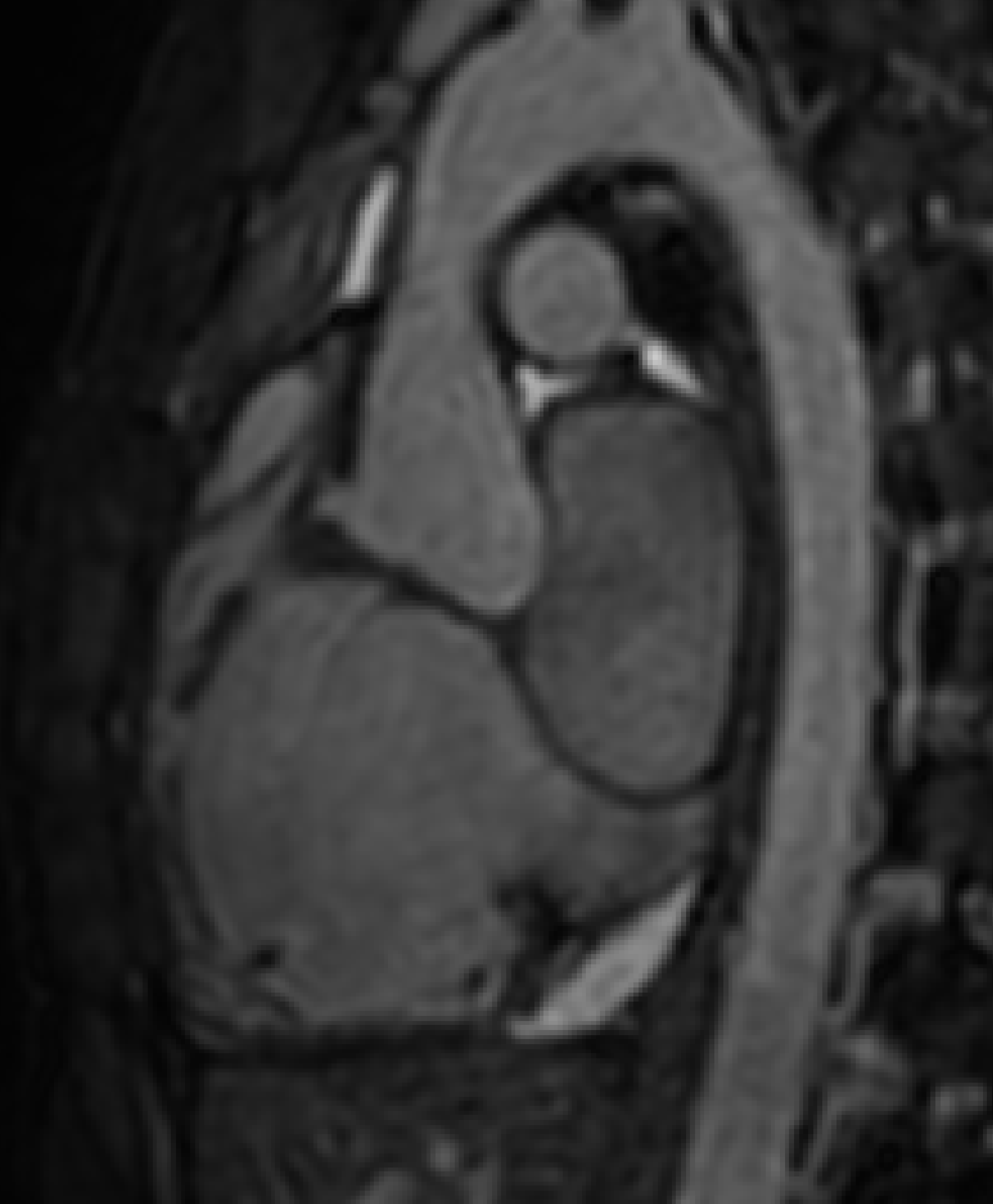}
\label{subfig:test18}
}
\hfill
\subfloat[]{
\includegraphics[width=0.12\textwidth]{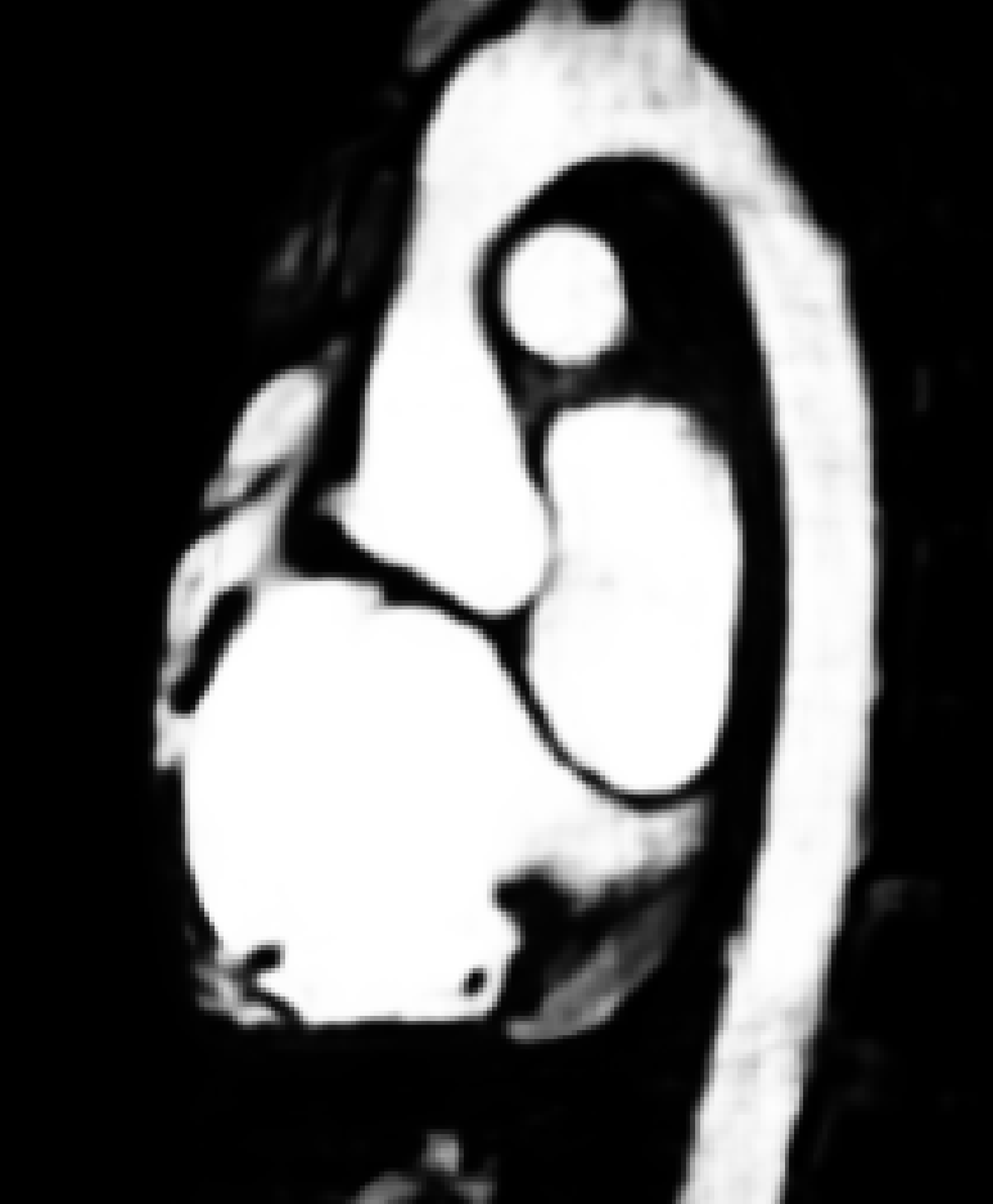}
\label{subfig:fold1}
}
\hfill
\subfloat[]{
\includegraphics[width=0.12\textwidth]{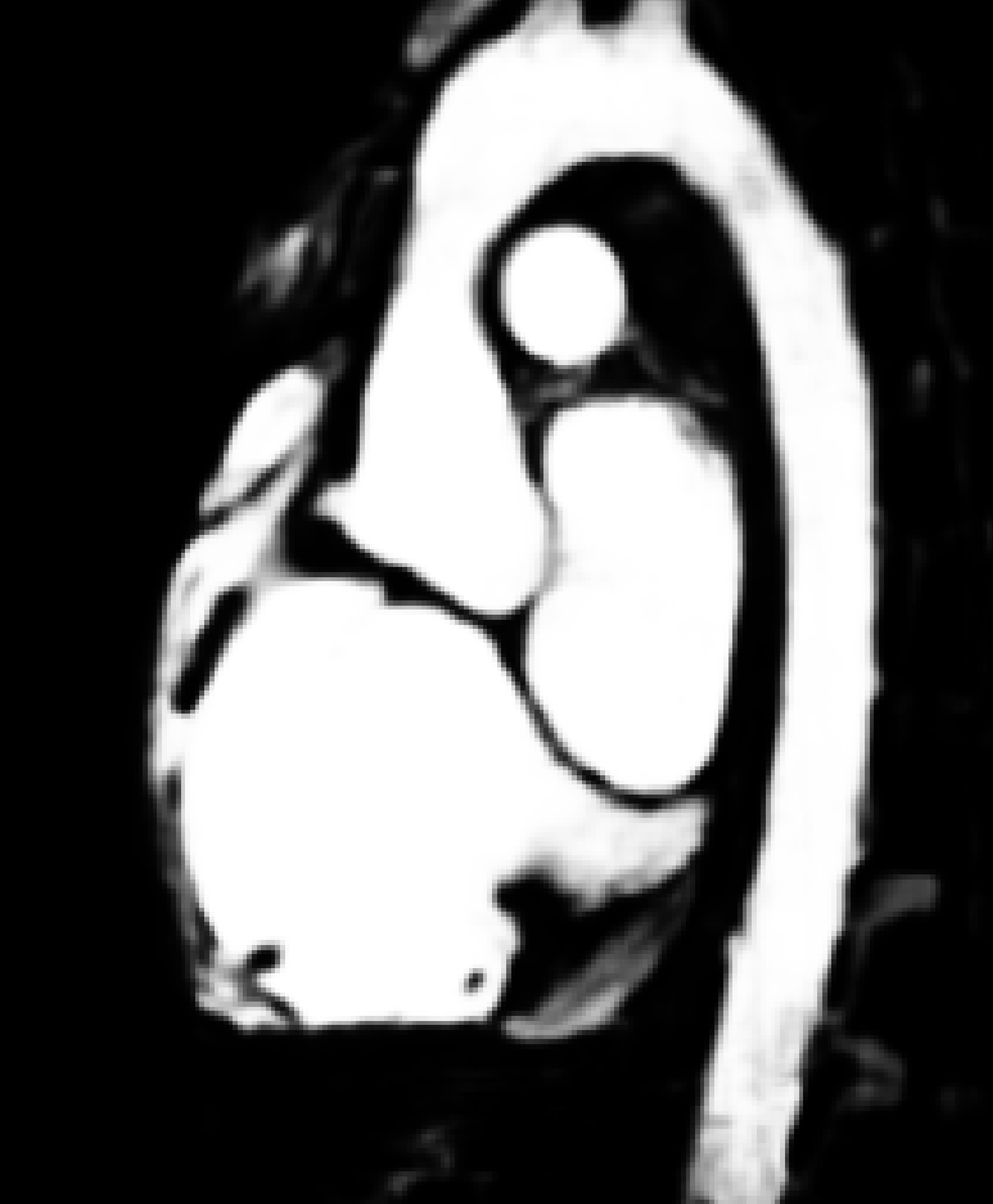}
\label{subfig:fold2}
}
\hfill
\subfloat[]{
\includegraphics[width=0.12\textwidth]{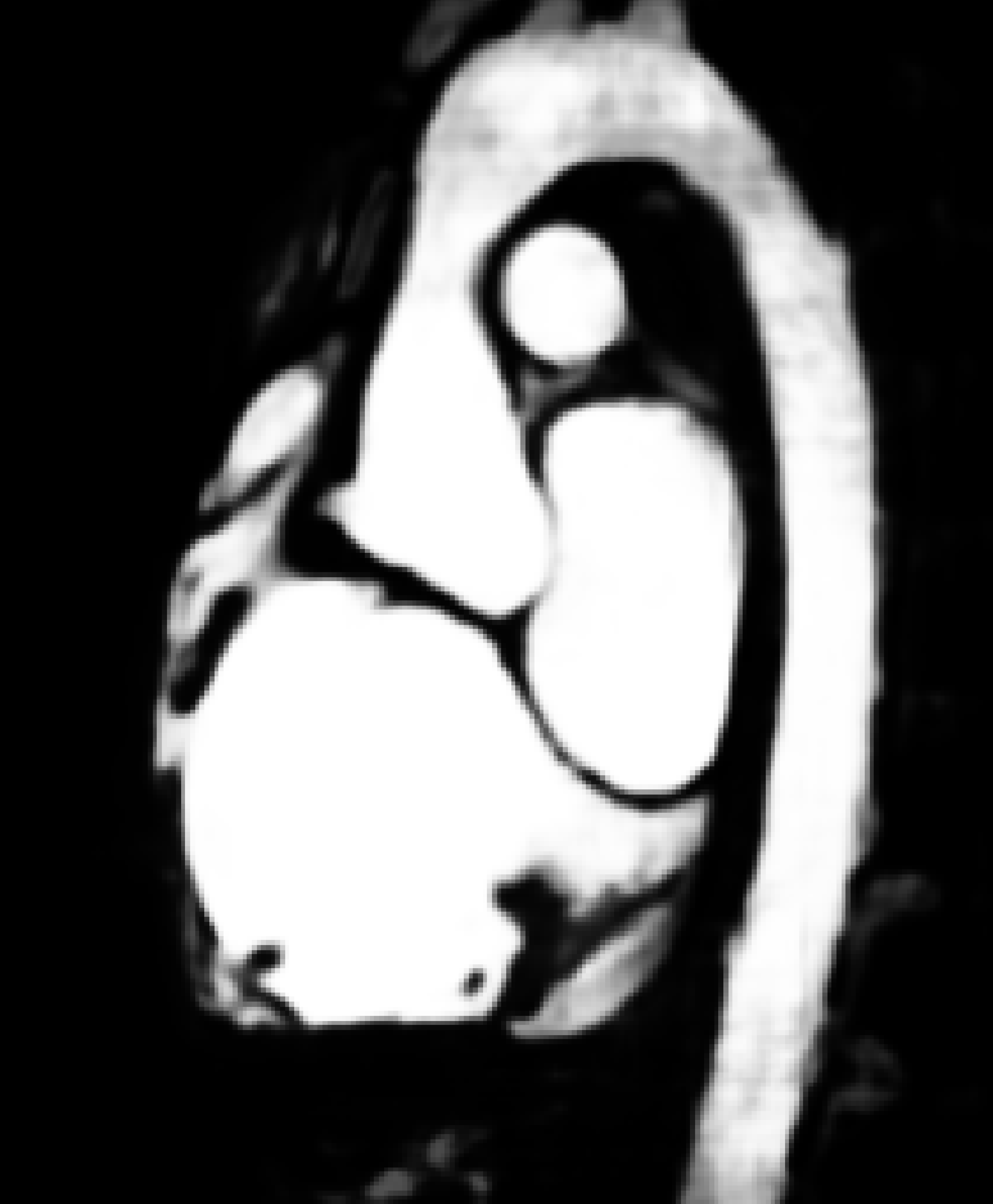}
\label{subfig:fold3}
}
\hfill
\subfloat[]{
\includegraphics[width=0.12\textwidth]{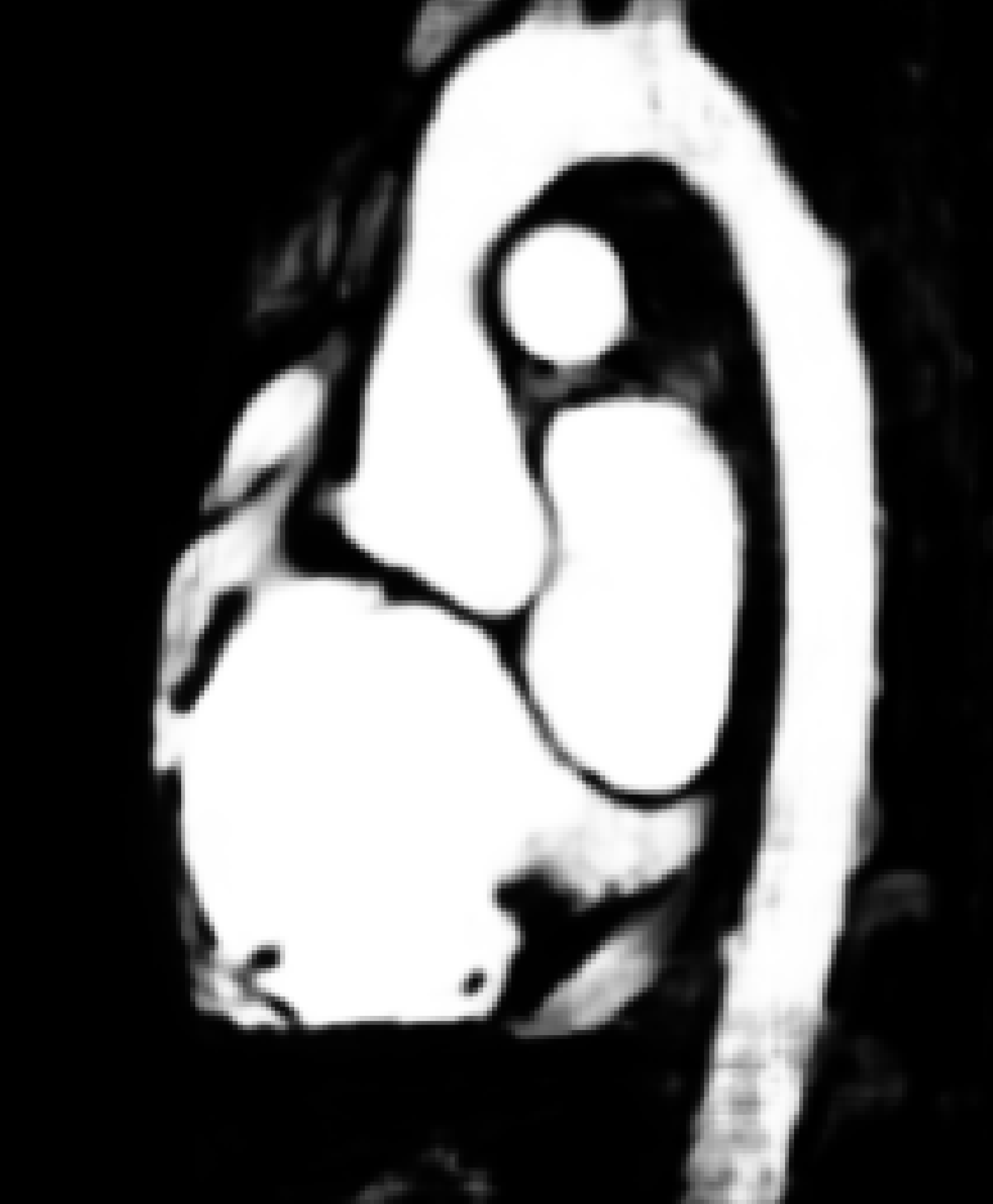}
\label{subfig:fold4}
}
\hfill
\subfloat[]{
\includegraphics[width=0.12\textwidth]{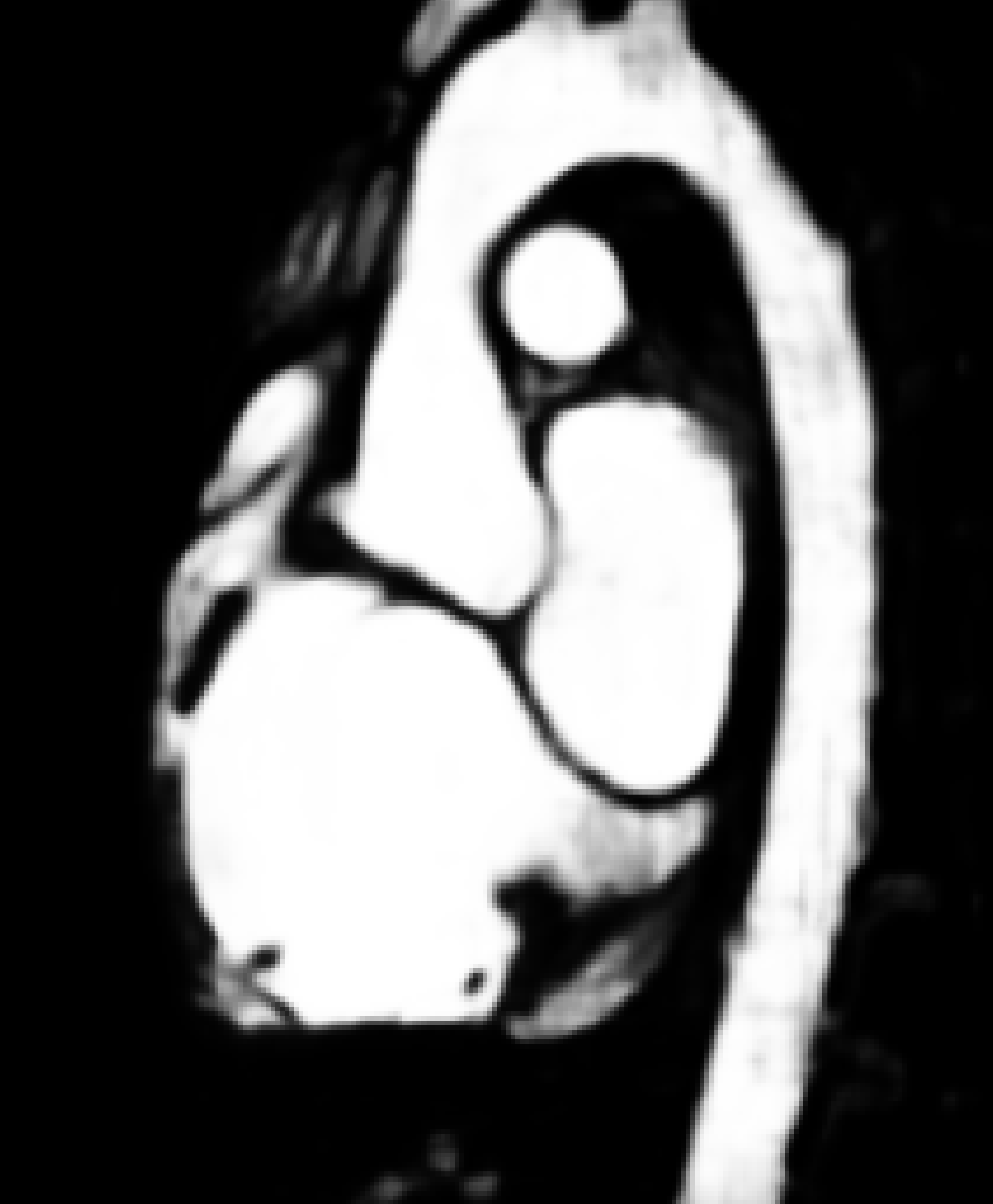}
\label{subfig:fold5}
}
\hfill
\subfloat[]{
\includegraphics[width=0.12\textwidth]{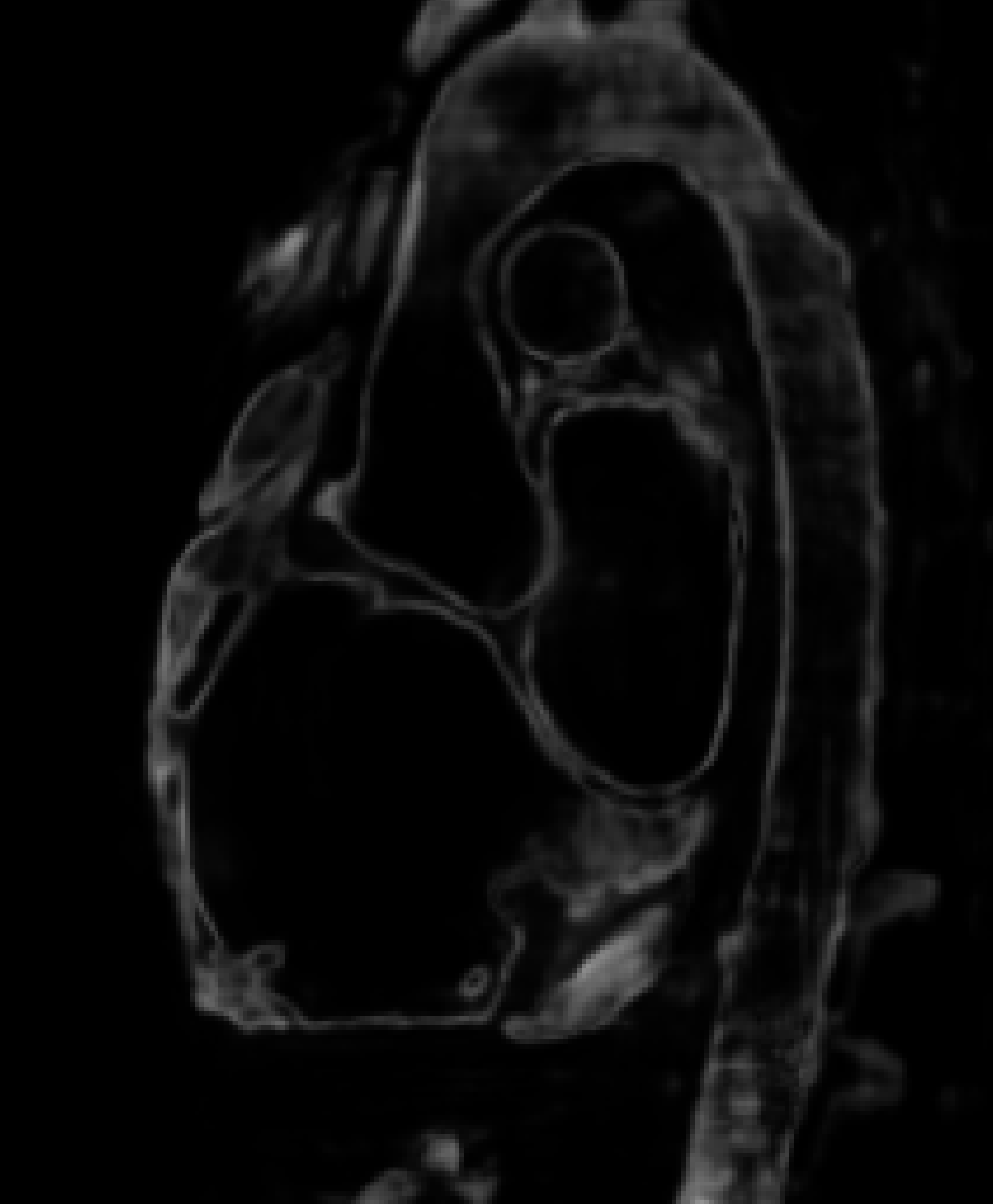}
\label{subfig:variance}
}
\hfill
\caption{\protect\subref{subfig:origslice} CMR image slice, Patient 18. \protect\subref{subfig:fold1}-\protect\subref{subfig:fold5} Probabilistic blood pool maps obtained using CNNs trained on five different folds during cross-validation. \protect\subref{subfig:variance} Standard deviation of probabilities predicted by CNNs.} 
\label{fig:stability}
\end{figure}

To compare the performance of a CNN with dilated convolutions and a CNN without dilated convolutions, segmentations were performed using an otherwise identical CNN architecture containing 72,643 trainable parameters. Fig. \ref{fig:withwithout} shows the obtained results.
By omitting dilation, the receptive field for each voxel was reduced. While local information was used in classification by the CNN without dilation, long-range information was not, and hence the network was much less specific than the network with dilation. 

To investigate whether overfitting may have ocurred in the five CNNs trained during cross-validation, we compared predictions made by these CNNs on an unseen image from the test set. Fig. \ref{fig:stability} shows this image and probabilistic blood pool maps obtained by the CNNs. Even though each CNN was trained with only eight training images, the variance among the five predictions (Fig. \ref{subfig:variance}) was generally low, with higher values at the edges of the blood pool. Hence, it is unlikely that overfitting to the training data in each fold occurred.

\section{Discussion and Conclusion}
We have presented a method for automatic segmentation of cardiovascular MR images in congenital heart disease using dilated convolutional neural networks. The method was able to accurately segment the myocardium and the blood pool without any expert intervention.

The current study showed that dilated convolution layers allow the combination of local structure and global context information with very few trainable parameters. Visual inspection of feature maps suggested that shallow layers enhanced local image features such as edges, and deeper layers distinguished between locally similar but globally different areas.
Our network used only 76,423 parameters, while comparable networks for medical image segmentation typically use more than 500,000 parameters \cite{Hava16}. This substantially reduces the risk of overfitting on the training data, which is particularly likely when training with very few scans as done in this study. In future work, we will investigate if the number of parameters can be further reduced without affecting performance, e.g. by reducing the number of output channels in each layer. The CNN was applied to full image slices to produce high-resolution output images, without downsampling of input or internal representations. We found that the method on average required only 41.5 seconds per scan, compared to 12.58 minutes in a recently published method for whole heart segmentation in cardiac MRI \cite{Zhua16}.

The method occasionally identified structures that were not included in the reference standard, but that are part of the blood pool, such as the distal sections of the descending aorta. It is unlikely that this will be problematic for the clinical purpose of segmentation of CMR in CHD patients. Dice indices for automatically obtained segmentations of the blood pool were in all patients higher than those of the myocardium. This is likely due to the lower image contrast between the myocardium and surrounding tissue, the more irregular shape of the myocardium and the difference in size between the myocardium and blood pool. 

For each voxel, the final label depended on three orthogonal patches centered at that voxel. The information in these patches was combined in a late stage, by averaging of the three probabilities provided by the CNN. 
In future work, the features extracted from the three orthogonal patches may be fused before classification. In addition, we will investigate dilated convolutions in 3D, which might allow us to fully leverage the volumetric information present in the image. However, hardware limitations currently force a trade-off between dimensionality and receptive field size, i.e. it would be infeasible to train a 3D dilated CNN with $131\times 131\times 131$ receptive fields. Therefore, we have here chosen to use a larger receptive field at the cost of reduced volumetric information.

\subsubsection{Acknowledgments}
We gratefully acknowledge the support of NVIDIA Corporation with the donation of the Tesla K40 GPU used for this research.

\bibliographystyle{splncs03}

\end{document}